\documentclass[journal]{IEEEtran}

\usepackage{url}
\usepackage{graphicx}
\usepackage{subfigure}
\usepackage{caption}
\usepackage{cite}
\usepackage[ruled, linesnumbered]{algorithm2e}
\usepackage{mathrsfs}
\usepackage{amsmath}
\usepackage{amssymb}
\usepackage{booktabs}
\usepackage{colortbl}
\usepackage{multirow}
\usepackage{amsthm}

\usepackage{float}
\usepackage{subfig}

\newtheorem{lemma}{Lemma}
\newtheorem{corollary}{Corollary}
\newtheorem{remark}{Remark}
\newtheorem{definition}{Definition}
\newtheorem{assumption}{Assumption}

\begin{document}

\title{{CyclicFL}: Efficient Federated Learning with Cyclic Model Pre-Training}


\author{
    \IEEEauthorblockN{Pengyu Zhang, Yingbo Zhou,  Ming Hu, Xian Wei, Mingsong Chen \thanks{* Corresponding author (mschen@sei.ecnu.edu.cn)}}
    
    \IEEEauthorblockA{MoE Engineering Research Center of SW/HW Co-Design Technology and Application, East China Normal University}
}


\maketitle

\begin{abstract}
Federated learning (FL) has been proposed to enable distributed learning on Artificial Intelligence Internet of Things
(AIoT) devices with guarantees of high-level data privacy.
Since random initial models in FL can easily result in unregulated  Stochastic Gradient Descent (SGD) processes, existing
FL methods greatly suffer from  both slow convergence and poor
accuracy, especially in non-IID scenarios. 
To address this problem, we propose a novel method named CyclicFL,  which can quickly derive effective initial models to
guide the SGD  processes, thus
improving the overall FL training performance. 
We formally
analyze the significance of data
consistency between the pre-training and training stages of CyclicFL, showing 
the limited Lipschitzness of loss for the pre-trained models by CyclicFL.
Moreover, we systematically prove that our method can achieve faster convergence speed under various convexity assumptions.
Unlike traditional centralized pre-training methods 
that require public proxy data,  CyclicFL pre-trains initial models 
on selected AIoT devices cyclically without exposing their local data. 
Therefore, they can be easily integrated into any security-critical FL methods. 
Comprehensive experimental results show that CyclicFL can
not only improve the maximum classification 
accuracy by up to $14.11\%$ but also significantly accelerate the overall FL 
training process.
\end{abstract}


\IEEEpeerreviewmaketitle

\section{Introduction}
\label{sec:intro}


\IEEEPARstart{A}{s} a vital
decentralized machine learning paradigm to bridge the Internet of Things (IoT) and Artificial Intelligence (AI),
Federated Learning (FL) \cite{fediot}
is promising to facilitate collaborative learning between Artificial Intelligence Internet
of Things (AIoT) \cite{iot_aiot} devices and cloud servers.
To avoid disclosure of privacy, in each 
FL communication round, all the involved AIoT devices first conduct local training and then upload their 
local gradients rather than private data 
to the cloud. The server updates the global model by aggregating received local gradients and dispatches the updated global model to devices for a new round of training.





Although FL is becoming increasingly popular in the design of security-critical AIoT applications, it suffers from slow convergence and poor 
accuracy, especially in non-IID (Independent and Identically Distributed) scenarios \cite{Zhu-et-al:noniidsurvey}. This is mainly 
because the inconsistencies of statistical properties among 
device data inevitably lead to learning divergences during the training of 
local models \cite{Wang-et-al:fedNova}, which strongly degrade
the performance of global model aggregation. 
Things become even worse when the initial device models are initialized 
randomly within a non-IID scenario. In this case, since devices tend to approach their local optimums during the FL training, it is hard for them to reach 
a consensus \cite{Li-et-al:moon}.

\begin{figure}[htbp]
    \centering
    \includegraphics[width=.9\linewidth]{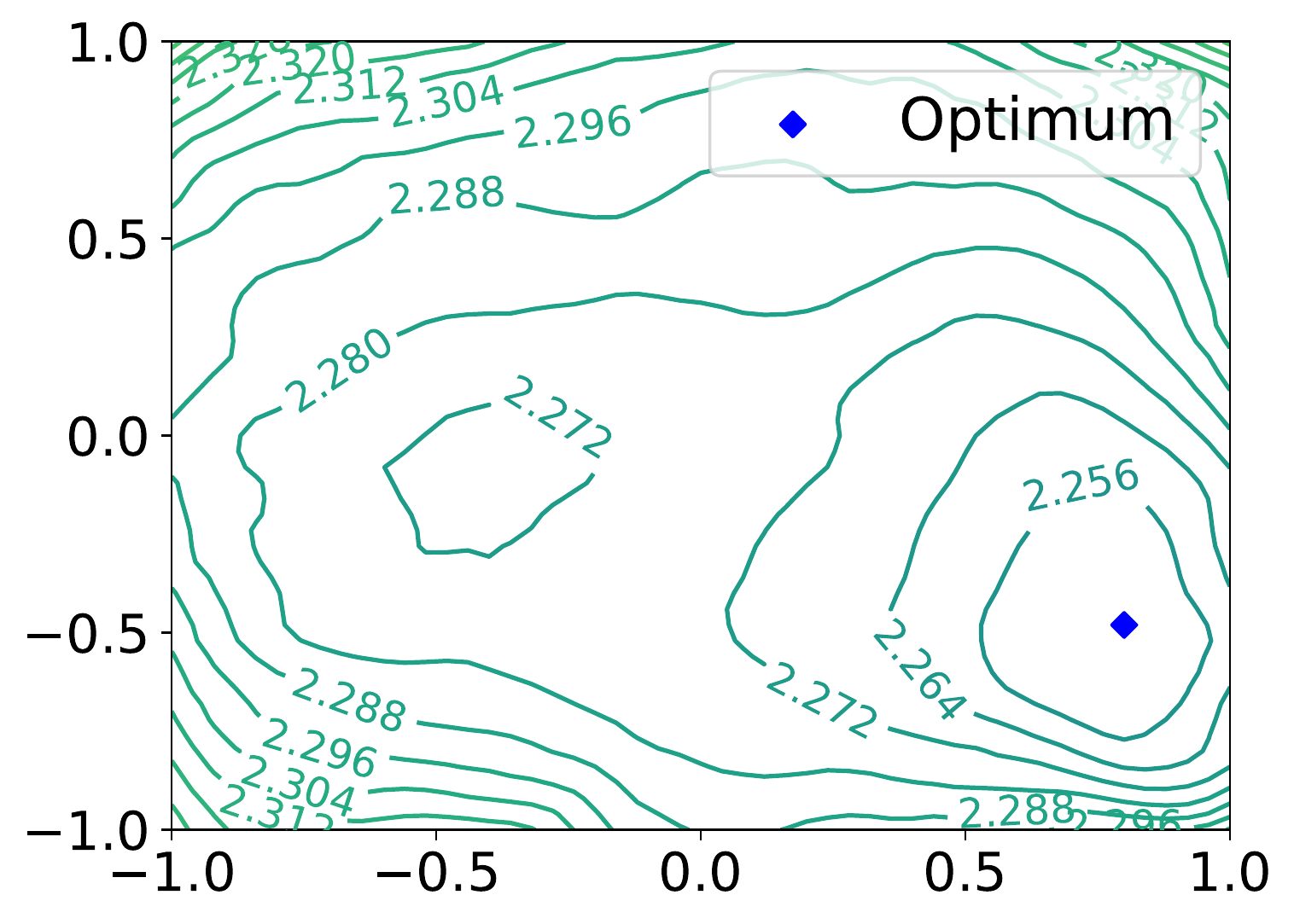}
    \caption{The loss landscape of a LeNet-5 initialized randomly. The training process is slowed due to the sharp landscape.}
    \label{failure}
\end{figure}

To improve the performance of FL training, various optimization methods have been proposed.  Unlike traditional post-processing methods 
(e.g., 
knowledge distillation-based methods~\cite{Hinton-et-al:distill}\cite{lu2023federatedmm})
, 
pre-training methods can implicitly capture transferable consensuses in the feature space to improve FL performance further.  
By establishing a better initialization for model parameters, pre-training is extraordinarily beneficial to downstream tasks by introducing affordable steps of fine-tuning~\cite{Howard-et-al:Fintune}. 
As an example shown in Figure \ref{failure},
without pre-training, the FL training starts with a randomly initialized model. 
The sharp loss landscape of the randomly initialized model causes a large variation in the loss function and slows the convergence according to \cite{Liu-et-al:transferunder}. 
Instead, our method creates a flatter loss landscape to stabilize and accelerate the downstream training~\cite{Liu-et-al:transferunder} and lead to faster convergence. 
Although pre-training can substantially
benefit
the FL training, existing pre-training
methods assume the existence 
of numerous
proxy data, which is not 
true in practice. Worse still, even if such  
proxy data is available, it is still
difficult to infer the data distributions of local AIoT devices from proxy data, thus restricting the application of pre-training in non-IID scenarios. 
In this paper, we propose a novel pre-training method named CyclicFL, which enables the quick generation of an effective initial model to accelerate the overall FL training process. 
Instead of using proxy data, CyclicFL pre-trains initial models using the local data of
cyclically
selected AIoT devices without exposing their
local privacy. Meanwhile, due to the 
significance of data consistency between the pre-training 
and training stages of CyclicFL, the benefits of pre-trained  models by CyclicFL on 
SGD processes can be guaranteed. In summary, this paper makes three major contributions: 
\begin{itemize}
    \item We propose a novel cyclic pre-training approach that can quickly derive initial models without using proxy data. The overall FL training performance can be dramatically improved by guiding the SGD processes on flat loss landscapes based on the pre-trained models.
    \item We show that our cyclic pre-training method guarantees data consistency between the pre-training and training stages. We present the convergence analysis, showing that our method exhibits a faster convergence speed.
    \item We implement CyclicFL on state-of-the-art FL methods, and conduct extensive experiments on well-known benchmarks to show the effectiveness of our method.
\end{itemize}


\section{Related Work} \label{sec:relate}
\noindent \textbf{Neural Network Initialization}.
As neural networks go deeper~\cite{Simonyan-Zisserman:verydeep}, the initialization of network parameters becomes important to prevent gradients from vanishing or explosion during the training procedure. 
So far, various initialization methods have been proposed. For example, Xavier-Initialization \cite{Glorot-Bengio:Xavier} normalized initial parameters to ensure the same variances of inputs and outputs under the assumptions of independence and identity between parameters. 
Since the Xavier-Initialization is only useful on saturating activation functions such as Sigmoid and Tanh~\cite{Nwankpa-et-al:sigtanh}, it was further extended to He-Initialization \cite{He-et-al:Heinit}, where ReLU~\cite{Nair-Hinton:rectified}, a broadly used non-saturating activation function, was taken into consideration.
Hu \textit{et al.}\cite{Hu-et-al:Hu2020Provable} provided theoretical proof that initializing model parameters with orthogonal groups accelerates convergence compared to that with Gaussian groups.
To adapt to more complex architectures used in vision tasks, e.g., ResNet (\cite{He-et-al:he2016deep}), MobileNet (\cite{Howard-et-al:howard2017mobilenets}), an initialization method from the perspective of training dynamics by estimating mutual information between two consecutive layers was proposed by \cite{Mao-et-al:neucam}.
A ``learning to initialize'' scheme ~\cite{Zhu-et-al:zhu2021gradinit} was proposed by formulating initialization on neural networks as an optimization problem.


\noindent\textbf{Pre-training in Federated Learning.}
Existing work has paid attention to the impact of pre-training in FL. \cite{Nguyen-et-al:wheretobegin} implemented comprehensive experiments and showed that starting federated training from pre-trained models not only improves the overall performances in FL but also alleviates the notorious local drift problems caused by data heterogeneity. Such approaches are crucial for AIoT scenarios since local devices are constrained by energy and communication overhead. 
To analyze the impact of pre-training on FL, 
\cite{Nguyen-et-al:wheretobegin} conducted comprehensive experiments on multiple FL benchmark datasets and pointed out that the impact of data heterogeneity~\cite{Li-et-al:fedproxy,Karimireddy-et-al:scaffold,Gao-et-al:gao2022feddc}, a notorious problem in FL, is much less severe if starting training from a pre-trained initialization. 
Another work \cite{Chen-et-al:prefl} emphasized that pre-trained models can boost model performances and close the gap to centralized training as it makes model aggregation more stable.
Furthermore, fine-tuning methods specifically designed for
vision-language tasks~\cite{Radford-et-al:CLIP} were proposed by \cite{Chen-et-al:fedtune} to explore
the benefits of using a pre-trained global model under the FL paradigm. 
The aforementioned works mainly concentrate on analyzing the impacts of pre-training on FL. Few of them provide solutions for obtaining a pre-trained model for neural networks given task-specific data \cite{Chen-et-al:prefl} or on which fine-tuning methods can induce more successful transfers from the pre-trained model to downstream FL tasks \cite{Chen-et-al:fedtune}. Most of the above methods leverage existing pre-trained models without proposing novel pre-training methods. Our solution, instead, aims to leverage task-specific data and provide a novel pre-training method to improve the performance of vanilla FL training for AIoT applications.

\section{Methodology} \label{sec:meth}
The goal of a general FL is to jointly learn a shared global model parameterized by $\mathbf{w}$ while keeping local data private. Considering $m$ local AIoT devices (clients) with private Non-IID dataset denoted as $\mathcal{D}_m=\{(\mathbf{x}_{m,j}, y_{m,j})\}_{j=1}^{N_m}$, we seek to solve the problem formulated as:
\vspace{-0.1in}
\begin{align}
    &\mathop{\min}\limits_{\mathbf{w}} f(\mathbf{w})= \sum\limits_{i=1}^{m}\frac{1}{N}\sum\limits_{j=1}^{N_i}\ell(\mathbf{w};(\mathbf{x}_{i,j},y_{i,j})).
    \vspace{-0.1in}
    \label{fedavg_obj}
\end{align}
Here, $m$ is the number of devices participating in local training in each round, $N_i$ denotes the number of instances on device $i$, and $N$ represents the total number of instances. $L_i(\mathbf{w}; \mathcal{D}_i)$ is the empirical loss on local dataset $\mathcal{D}_i$ consists of the loss function $\ell(\cdot)$. The neural network parameter $\mathbf{w}$ is assumed to be identical across local devices and the server under a homogeneous architecture setting.

\begin{figure*}[h]
\centering  
\includegraphics[width=0.9\linewidth]{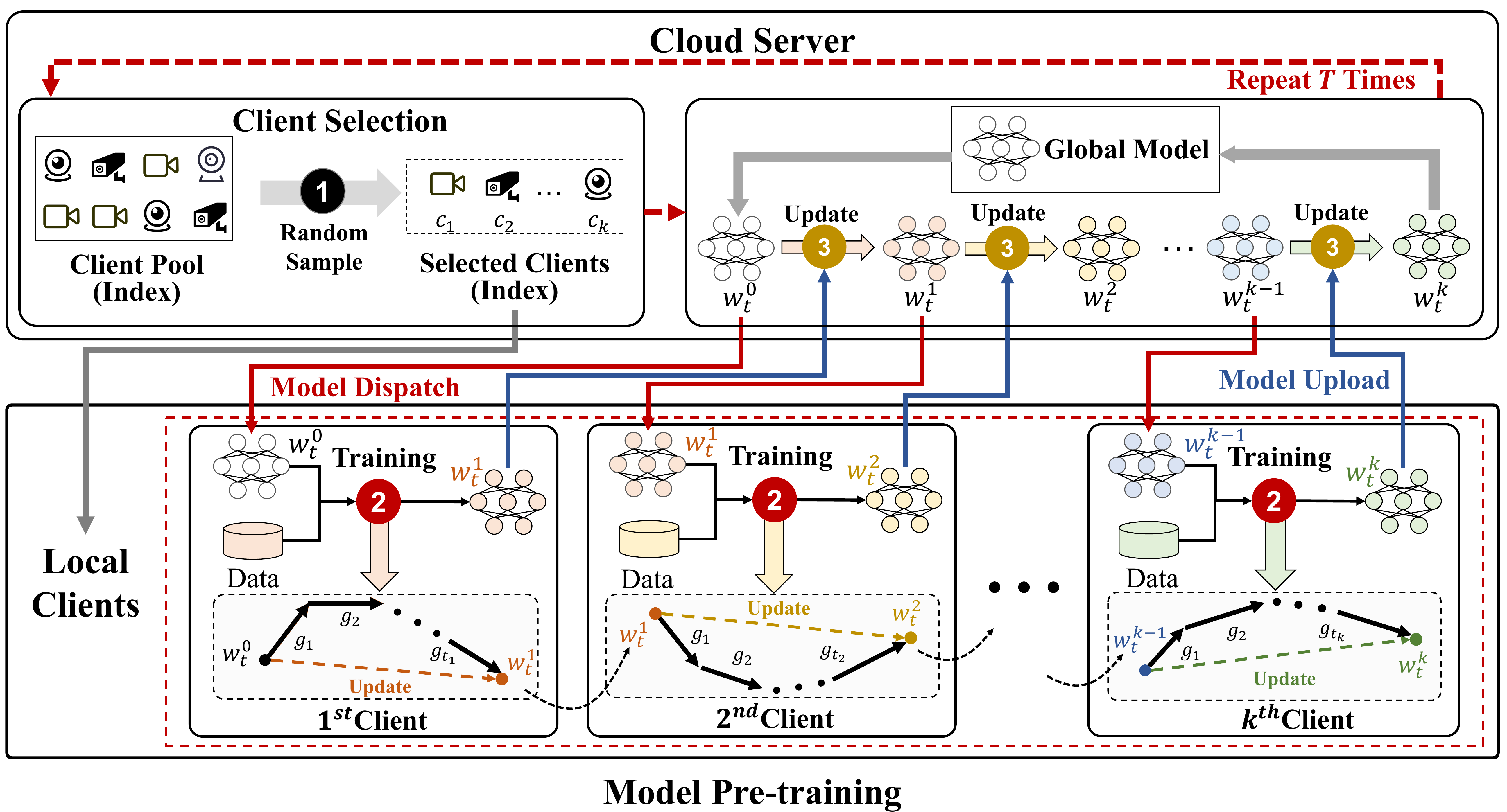} 
\caption{Workflow of our pre-training approach.}
\label{workflow}
\end{figure*}

Instead of initializing the federated training based on a randomly initialized global model $\mathbf{w_{rg}}$, we aim to find a well-initialized $\mathbf{w_{wg}}$ that can improve the performance and accelerate the optimization objective defined in Eq.~\ref{fedavg_obj}.
Figure~\ref{workflow} shows the main workflow of our cyclic pre-training method. The entire workflow is built on two alternate steps, i.e., the selection of a group of local devices and the local training. 
We denote every two consecutive steps as one round.
The server on the cloud first performs a group selection step by randomly sampling a group of local AIoT devices, on which a series of local training will be conducted. 
In the meantime, the server creates the randomly initialized model. Note that the server only randomly initializes the model at the beginning of the first round. For simplicity, we use P1 and P2 to denote the cyclic pre-training phase and FL training phase. 
Note that our method differs from the sequential training approach proposed by \cite{li2023convergence}. Unlike directly applying sequential training, we introduce a pre-training phase to find a well-initialed model that facilitates faster convergence and enhances overall performance.

Algorithm~\ref{alg:aggregation} describes the implementation of cyclic pre-training. 
At the beginning of cyclic pre-training (Line $1$), the cloud server randomly initializes a model parameterized by $\mathbf{w_{rg}}$. Lines $2-12$ implement $T$ rounds of cyclic pre-training in total. Line $3$ describes the $RandomSample$ process, where the cloud server randomly samples a set ($S_t$) of local devices to participate in the local training in this round. In line $4$, the constant $\mathcal{K}_{P1}$ represents the number of devices in set $S_t$. After the sampling of local devices is completed, Lines $5-10$ describe the process of local training. Line $5$ represents the outer loop, in which the cloud server broadcasts the global model to the $i^{th}$ device to perform the local training. Lines $6-8$ represent the inner loop, where the $i^{th}$ device performs $t_i$ local training steps (i.e., $\mathbf{g}_1,\cdots, \mathbf{g}_{t_i}$) with its private data. At the end of each inner loop, the local device sends the trained model back to the server. The server further transmits the trained model to the first device in the next round to initiate a new outer loop. Such server-to-local and local-to-server alternate steps are described in Line $9$. Line $11$ shows the procedure that initiates a new round of cyclic pre-training. Line $13$ returns a pre-trained global model parameterized by $\mathbf{w_{wg}}$.
We use the notation $\mathbf{w}_t^{i,j}$ to denote the model parameters, where $t$, $i$, $j$ represent model parameters in the $j^{th}$ training step, the $i^{th}$ local device, and the $t^{th}$ round, respectively.
We consider training a neural network parameterized by $\mathbf{w}_t^0$ on a set of data denoted by $S_t$ in the $t^{th}$ round, where $S_t = \{D_1^t, D_2^t,\dots,D_k^t\}$.
In round $t$, $k$ datasets that are used to train the model are listed in a fixed order, i.e., from $D_1^t$ to $D_k^t$, yielding an evolution of neural network parameters from $\mathbf{w}_t^0$ to $\mathbf{w}_t^k$. On the $i^{th}$ local device in the $t^{th}$ round, $\mathbf{w}_t^i$ is obtained by fitting model $\mathbf{w}_t^{i-1}$ on dataset $D_i^t$ with $t_i$ steps of optimizations (SGD). 
The cyclic pre-training is formulated as \scalebox{1.0}{$\mathop{\min}\limits_{\mathbf{w}}\mathcal{F}(\mathbf{w})= \sum\limits_{t=1}^{T}\sum\limits_{j=1}^{k}\ell(\mathbf{w}; \mathcal{D}_j^t(n))$},
where $\mathcal{F}(\mathbf{w})$ is the objective function of cyclic pre-training, and $T$ represents the number of rounds cyclic pre-training consumes. 
The Local data of the $j^{th}$ device in the $t^{th}$ round is denoted as $\mathcal{D}_j^t(n) = \{(\mathbf{x}_{i,1}, y_{i,1}),(\mathbf{x}_{i,2}, y_{i,2}),\dots,(\mathbf{x}_{j,n}, y_{j,n})\}$. The loss function $\ell$ is determined by specific tasks, e.g., cross-entropy loss for multi-classification tasks.

\vspace{-4mm}
\begin{algorithm}[ht]
\caption{Our Cyclic Pre-training Algorithm}
\label{alg:aggregation}
\textbf{Input}: 
i) $\mathbf{w_{rg}}$, a randomly initialized global model; 
ii) $S$, a pool of all devices; 
iii) $T$, \# of cyclic pre-training rounds; 
iv) $\mathcal{K}_{P1}$, \# of devices in each round.     
\\
\textbf{Output}: 
 $\mathbf{w_{wg}}$, a well-initialized global model. \\
\textbf{Cyclic Pre-training:}

\SetAlgoLined
\KwIn{$\mathbf{w_{rg}}$, $S$, $T$, $\mathcal{K}_{P1}$}
\KwOut{$\mathbf{w_{wg}}$}
$\mathbf{w}_1^{1,0} \leftarrow \mathbf{w_{rg}}$\;
\For{$t=1,\dots,T$}{
    $S_t \leftarrow RandomSample(S, \mathcal{K}_{P1})$\;
    $k \leftarrow$ Number of elements in $S_t$\;
    \For{$i = 1,\dots, k$}{
        \For{$j = 1,\dots, t_i$}{
            $\mathbf{w}_t^{i,j} \leftarrow \mathbf{w}_t^{i,j-1} - \mathbf{g}_j$\;
        }
        $\mathbf{w}_t^{(i+1),0} \leftarrow \mathbf{w}_t^{i,t_i}$\;
    }
    $\mathbf{w}_{t+1}^{1,0} \leftarrow \mathbf{w}_{t}^{k,t_{k}}$\;
}
$\mathbf{w_{wg}} \leftarrow \mathbf{w}_{T}^{k,t_{k}}$\;
\Return{$\mathbf{w_{wg}}$}
\end{algorithm}

\section{Theoretical Analysis}\label{sec:theory}
\subsection{Data consistency on Lipschitzness of loss}
In Lemma~\ref{lossbasin}, we show that our pre-training method
ensures the data consistency between the pre-training data and downstream FL data, which is crucial to the performance of pre-training. In contrast, breaking such consistency directly causes the failure of a pre-training process.

\begin{definition}\label{def_data_consist}
Let $\lambda_P$ and $\lambda_Q$ represent the smallest eigenvalues of $\mathbf{H}_{P }^{\infty}$ and $\mathbf{H}_{Q}^{\infty}$, respectively. Let $\mathbb{I}$ denote the indicator function. The Gram matrix of input data $\mathbf{H}_{P }^{\infty}$ and $\mathbf{H}_{Q}^{\infty}$ are presented as blew:
\begin{small}
\begin{align}
\nonumber\mathbf{H}_{P, i j}^{\infty} & = \mathbb{E}_{\mathbf{w} \sim \mathcal{N}(\mathbf{0}, \mathbf{I})}\left[\mathbf{x}_{P, i}^{\top} \mathbf{x}_{P, j} \mathbb{I}\left\{\mathbf{w}^{\top} \mathbf{x}_{P, i} \geq 0, \mathbf{w}^{\top} \mathbf{x}_{P, j} \geq 0\right\}\right] \\
&= \frac{\mathbf{x}_{P, i}^{\top} \mathbf{x}_{P, j}\left(\pi-\arccos \left(\mathbf{x}_{P, i}^{\top} \mathbf{x}_{P, j}\right)\right)}{2 \pi},
\label{hp}
\end{align}
\vspace{-0.3in}

\begin{align}
\nonumber\mathbf{H}_{Q, i j}^{\infty} & = \mathbb{E}_{\mathbf{w} \sim \mathcal{N}(\mathbf{0}, \mathbf{I})}\left[\mathbf{x}_{Q, i}^{\top} \mathbf{x}_{Q, j} \mathbb{I}\left\{\mathbf{w}^{\top} \mathbf{x}_{Q, i} \geq 0, \mathbf{w}^{\top} \mathbf{x}_{Q, j} \geq 0\right\}\right]\\
&= \frac{\mathbf{x}_{Q, i}^{\top} \mathbf{x}_{Q, j}\left(\pi-\arccos \left(\mathbf{x}_{Q, i}^{\top} \mathbf{x}_{Q, j}\right)\right)}{2 \pi},
\label{hq}
\end{align}

\vspace{-0.2in}

\begin{align}
\nonumber\mathbf{H}_{P Q, i j}^{\infty} & = \mathbb{E}_{\mathbf{w} \sim \mathcal{N}(\mathbf{0}, \mathbf{I})}\left[\mathbf{x}_{P, i}^{\top} \mathbf{x}_{Q, j} \mathbb{I}\left\{\mathbf{w}^{\top} \mathbf{x}_{P, i} \geq 0, \mathbf{w}^{\top} \mathbf{x}_{Q, j} \geq 0\right\}\right] \\
&= \frac{\mathbf{x}_{P, i}^{\top} \mathbf{x}_{Q, j}\left(\pi-\arccos \left(\mathbf{x}_{P, i}^{\top} \mathbf{x}_{Q, j}\right)\right)}{2 \pi}.
\label{hpq}
\end{align}
\end{small}

$\mathbf{H}_{P, ij}^{\infty}$ and $\mathbf{H}_{Q, ij}^{\infty}$ represent the Gram matrix of input data in dataset P and dataset Q, respectively. $\mathbf{H}_{P Q, i j}^{\infty}$ measures the relations between P and Q. Notation $\mathbf{w}$ is the model parameter, and $\mathbb{I}$ is the indicator function in Equations~\ref{hp},~\ref{hq}, and~\ref{hpq}. The number of neurons in the given network is defined as m. $\mathbf{y}_{P\rightarrow Q}$ is derived from $(\mathbf{H}_{P Q}^{\infty})^T(\mathbf{H}_{P}^{\infty})^{-1}\mathbf{y}_P$. 


\end{definition}

\begin{lemma}
(\textbf{The impact of transferred features on the Lipschitzness of the loss}) Let Q represent the target dataset, P represent the pre-training dataset, and $\operatorname{poly}$ represent a polynomial.  For a two-layer network with a large number (m) of hidden neurons, if m $\geq \operatorname{poly}\left(n_{P}, n_{Q}, \delta^{-1}, \lambda_{P}^{-1}, \gamma^{-1}\right)$, $\gamma=O\left((n_{P}^{2} n_{Q}^{\frac{1}{2}}) / (\lambda_{P}^{2} \delta)\right)$, with a probability no less than $1-\delta$ over the random initialization, the Lipschitzness of loss is formulated as follows \cite{Liu-et-al:transferunder}:
\label{lossbasin}
\begin{align} 
\footnotesize
\nonumber\left\|\frac{\partial L(\mathbf{w}(P))}{\partial \mathbf{X}^{1}}\right\|^{2} & \approx\left\|\frac{\partial L(\mathbf{w}(0))}{\partial \mathbf{X}^{1}}\right\|^{2}-\mathbf{y}_{Q}^{\top} \mathbf{y}_{Q} \\
\nonumber&+\|\mathbf{y}_{Q}-\mathbf{y}_{P \rightarrow Q}\|_2^2 + \mathcal{C}
\end{align}
where $0 < \gamma \leq 1$ controls the magnitude of initial network parameters $\mathbf{w}(0)$. $\mathcal{C}$ denotes $\frac{\operatorname{poly}\left(n_{P}, n_{Q}, \delta^{-1}, \lambda_{P}^{-1}, \gamma^{-1}\right)}{m^{\frac{1}{4}}}$.
\end{lemma}

Constants $n_p$ and $n_Q$ represent the number of training samples in the pre-training and target datasets, respectively. The corresponding labels are denoted as $\mathbf{y}_p$ and $\mathbf{y}_Q$, respectively. The matrix $\mathbf{X}^1$ denotes the activations of the neural network in the target dataset. 
Lemma~\ref{lossbasin} shows that the Lipschitzness of loss is controlled by the similarity between tasks in both input and labels. 
The Lipschitzness depicts the continuity of a given loss function, therefore showing the flatness of the loss landscape.
If the label used in pre-training is similar to the target task label, the difference between labels is smaller. Thus, the term $\|\mathbf{y}_{Q}-\mathbf{y}_{P\rightarrow Q}\|_2^2$ decreases and the Lipschitzness of the loss function increases. Combining the theoretical property of loss basins with observations from \cite{Chen-et-al:prefl}, we conclude that since our method utilizes partial task data as the pre-training data, the data consistency is greatly enhanced. Therefore, our method creates a flat loss basin and can successfully generate pre-training models tailored for downstream FL training tasks.



\subsection{Convergence analysis}
We investigate
the convergence rate of our method based on the assumptions and analysis made for Sequential Federated Learning (SFL) and Parallel Federated Learning (PFL) \cite{li2023convergence}. 
\begin{assumption}\label{assum1}
Each local objective function $\ell_i(\cdot)$ is $L$-smooth, $i\in\{1,2,\ldots,m\}$ i.e., there exists a constant $L>0$ such that
\begin{align}
    \left\|\nabla \ell_i(\mathbf{w})-\nabla \ell_i(\mathbf{z})\right\|\leq L\left\|\mathbf{w}-\mathbf{z}\right\|\textit{for all }\mathbf{w},\mathbf{z}\in\mathbb{R}^d.
\end{align}
\end{assumption}
\begin{assumption}\label{assum2}
The variance of the stochastic gradient at device $i$ is bounded:
\begin{align}
    \left.\mathbb{E}_{\xi\sim\mathcal{D}_{i}}\left[\left\|\nabla \ell_{i}(\mathbf{w};\xi)-\nabla \mathcal{L}_{i}(\mathbf{w})\right\|^{2}\right|\mathbf{w}\right]\leq\sigma^{2},\;\forall i\in\{1,2,\ldots,m\}.
\end{align}
\end{assumption}
\begin{assumption}\label{assum3}
There exist constants $\beta^2$ and $\zeta^2$ such that
\begin{align}
    \frac{1}{m} \sum_{i = 1}^{m}\left\|\nabla \ell_{i}(\mathbf{w})-\nabla \mathcal{L}(\mathbf{w})\right\|^{2} \leq \beta^{2}\|\nabla \mathcal{L}(\mathbf{w})\|^{2}+\zeta^{2}.
\end{align}
\end{assumption}

\begin{assumption} \label{assum4}
There exist a constant $\zeta_*^2$ such that
\begin{align}
    \nonumber\frac{1}{m}\sum_{i=1}^m\left\|\nabla L_i(\mathbf{w}^*)\right\|^2=\zeta_*^2
\end{align}
where $\mathbf{w}^*$ is the global minimizer.
\end{assumption}

\begin{remark}\label{remark}
Considering that performing SFL and PFL individually for $T$ steps with non-convex assumptions. The gradient norm of SFL denoted as $\mathbb{E}\left[\left\|\nabla \ell\left(\mathbf{w}^{(T)}\right)\right\|^{2}\right]$ is smaller than that of PFL. Therefore, the initial state of PFL $\mathbf{w}^{(t)}_{p}$ inherited from SFL $\mathbf{w}^{(t)}_{s}$ is no worse than the random initialization.
\end{remark}

\begin{definition}\label{def}
Let $S_c$ and $S_p$ represent the number of devices in each round in P1 and P2, respectively. Let $K_c$ and $S_p$ denote the local training steps in P1 and P2. $A_c$ represents $\mathcal{L}(\mathbf{w}^{(0)})-\mathcal{L}(\mathbf{w}^{T_{c}})$ and $A_p$ represents $\mathcal{L}(\mathbf{w}^{T_{c}})-\mathcal{L}(\mathbf{w}^{T_{total}})$.
\end{definition}

\begin{corollary}\label{corollary1}
By choosing appropriate learning rates, we can obtain the convergence bounds for CyclicFL as follows:

\textbf{Strongly convex:}
Let $\mathbb{E}$ represent 
$\mathbb{E}\left[\mathcal{L}(\bar{\mathbf{w}}^{T})-\mathcal{L}(\mathbf{w}^{*})\right]$.
Let $D_c$ represent $\|\mathbf{w}^0 - \mathbf{w}^{T_c} \|$ and $D_p$ represent $\|\mathbf{w}^{T_c} - \mathbf{w}^{T_{total}} \|$.
Under Assumptions~\ref{assum1},\ref{assum2}, \ref{assum4}, and Definition~\ref{def}, there exists a constant effective learning rate $\frac{\mu}{T}\leq\tilde{\eta}\leq\frac{1}{6L}$ that the convergence rate in a strongly convex scenario is defined as

\noindent for $T\leq T_c$ ,
\begin{align}
\nonumber\mathbb{E}=\mathcal{O}\bigg(&\mu D_c^{2}\exp\left(-\frac{\mu R}{12L}\right)+\frac{\sigma^{2}}{\mu S_cK_cT} \\
&+\frac{(m-S_c)\zeta_{*}^{2}}{\mu S_cT(m-1)}+
\frac{L\sigma^{2}}{\mu^{2}S_cK_cT^{2}}+\frac{L\zeta_{*}^{2}}{\mu^{2}S_cT^{2}}\bigg)
;
\end{align}

\noindent for $T_c<T<T_{total}$ ,
\begin{align}
\nonumber\mathbb{E}=\mathcal{O}\bigg(&\mu D_p^{2}\exp\left(-\frac{\mu T}{12L}\right)+\frac{\sigma^{2}}{\mu S_pK_pT} \\
&+\frac{(m-S_p)\zeta_{*}^{2}}{\mu S_pT(m-1)}+\frac{L\sigma^{2}}{\mu^{2}K_pT^{2}}+\frac{L\zeta_{*}^{2}}{\mu^{2}T^{2}}\bigg).
\end{align}

\textbf{General convex:}
Let $\mathbb{E}$ represent 
$\mathbb{E}\left[\mathcal{L}(\bar{\mathbf{w}}^{T})-\mathcal{L}(\mathbf{w}^{*})\right]$. Let $D_c$ represent $\|\mathbf{w}^0 - \mathbf{w}^{T_c} \|$ and $D_p$ represent $\|\mathbf{w}^{T_c} - \mathbf{w}^{T_{total}} \|$
Under Assumptions~\ref{assum1},\ref{assum2}, \ref{assum4}, and \textbf{Definition~\ref{def}}, there exists a constant effective learning rate $\tilde{\eta}\leq\frac{1}{6L}$ that the convergence rate in a general convex scenario is defined as

\noindent for $T\leq T_c$ ,
\begin{align}
\nonumber\mathbb{E}=
\mathcal{O}\bigg(&\frac{\sigma D_c}{\sqrt{S_cK_c}}+\sqrt{1-\frac{S_c}{M_c}}\cdot\frac{\zeta_{*}D_c}{\sqrt{S_cT}} \\
&+\frac{\left(L\sigma^{2}D_c^{4}\right)^{1/3}}{(S_cK_c)^{1/3}T^{2/3}}+\frac{\left(L\zeta_{*}^{2}D_c^{4}\right)^{1/3}}{S_c^{1/3}T^{2/3}}+\frac{LD_c^{2}}{T}\bigg)
;
\end{align}

\noindent for $T_c<T<T_{total}$ ,
\begin{align}
\scriptsize
\nonumber\mathbb{E}=\mathcal{O}\bigg(&\frac{\sigma D_p}{\sqrt{S_pK_pT}}+\sqrt{1-\frac{S_p}{m}}\cdot\frac{\zeta_{*}D_p}{\sqrt{S_pT}} \\
+&\frac{\left(L\sigma^{2}D_p^{4}\right)^{1/3}}{K_p^{1/3}T^{2/3}}+\frac{\left(L\zeta_{*}^{2}D_p^{4}\right)^{1/3}}{T^{2/3}}+\frac{LD_p^{2}}{T}\bigg).
\end{align}

\textbf{Non-convex:}
Let $\mathbb{E}$ represent $\min _{0 \leq t \leq T} \mathbb{E}\left[\left\|\nabla \ell\left(\mathbf{w}^{(t)}\right)\right\|^{2}\right]$. Under Assumptions~\ref{assum1},\ref{assum2}, \ref{assum3}, Remark~\ref{remark} and Definition~\ref{def}, there exists a constant effective learning rate $\tilde{\eta}\leq\frac{1}{6L(\beta+1)}$ that the convergence rate in a non-convex scenario is defined as:

\noindent for $T\leq T_c$ ,
\begin{align}
\scriptsize
\nonumber\mathbb{E}=\mathcal{O}\left(\frac{\left(L \sigma^{2} A_c\right)^{1 / 2}}{\sqrt{S_c K_c T}}+\frac{\left(L^{2} \sigma^{2} A_c^{2}\right)^{1 / 3}}{(S_c K_c)^{1 / 3} T^{2 / 3}}+\frac{\left(L^{2} \zeta^{2} A_c^{2}\right)^{1 / 3}}{S_c^{1 / 3} T^{2 / 3}}\right);
\end{align}

\noindent for $T_c<T<T_{total}$ ,
\begin{align}
\scriptsize
\nonumber\mathbb{E}=\mathcal{O}\left(\frac{\left(L\sigma^2A_p\right)^{1/2}}{\sqrt{S_pK_pT}}+\frac{\left(L^2\sigma^2A_p^2\right)^{1/3}}{K_p^{1/3}T^{2/3}}+ \frac{\left(L^2\zeta^2A_p^2\right)^{1/3}}{T^{2/3}}\right).
\end{align}

\end{corollary}

\noindent Based on Corollary~\ref{corollary1}, the overall convergence rate is determined by the cyclic pre-training stage and federated training stage. The convergence rate of cyclic pre-training has an advantage of $1/S_c$ on the second and third terms compared to the federated training. 

\begin{table*}
\caption{Comparison of test accuracy. 
}
\label{exp_randomseeed}
\centering
\scalebox{1.0}{
\begin{tabular}{c|l|l|c|ccccc} 
\hline
\multirow{2}{*}{Model} & \multicolumn{2}{c|}{\multirow{2}{*}{Dataset}} & \multicolumn{1}{l|}{\multirow{2}{*}{\begin{tabular}[c]{@{}l@{}}Non-IID \\ Setting ($\beta$)\end{tabular}}} & \multicolumn{5}{c}{Accuracy (\%)} \\ 
\cline{5-9}
 & \multicolumn{2}{c|}{} & \multicolumn{1}{l|}{} & FedAvg & FedProx & SCAFFOLD & Moon & Cyclic + FedAvg \\ 
\hline
CNN-FEMNIST & \multicolumn{2}{c|}{FEMNIST} & - & $81.44\pm1.26$  & $81.44\pm1.11$  & $\mathbf{83.48\pm0.48}$  & $81.46\pm0.90$  & $82.95\pm0.40$  \\ 
\hline
\multirow{3}{*}{CNN-Fashion-MNIST} & \multicolumn{2}{c|}{\multirow{3}{*}{Fashion-MNIST}} & 0.1 & $76.18\pm1.43$ & $76.11\pm2.01$ & $78.50\pm2.21$ & $76.28\pm1.34$ & $\mathbf{83.11\pm1.09}$ \\
 & \multicolumn{2}{c|}{} & 0.5 & $84.02\pm2.16$ & $84.01\pm2.19$ & $85.37\pm0.93$ & $83.94\pm1.97$ & $\mathbf{86.83\pm0.62}$ \\
 & \multicolumn{2}{c|}{} & 1.0 & $86.28\pm1.46$ & $86.18\pm1.42$ & $86.11\pm1.39$ & $85.89\pm1.50$ & $\mathbf{88.46\pm0.23}$ \\ 
\hline
\multirow{3}{*}{LeNet-5} & \multicolumn{2}{c|}{\multirow{3}{*}{CIFAR-10}} & 0.1 & $49.05\pm0.84$ & $49.00\pm0.23$ & $\mathbf{53.40\pm0.86}$ & $48.90\pm 0.45$ & $53.38\pm0.71$ \\
 & \multicolumn{2}{c|}{} & 0.5 & $53.22\pm0.60$ & $52.88\pm1.38$ & $\mathbf{60.14\pm1.53}$ & $52.27\pm0.27$ & $57.91\pm0.04$ \\
 & \multicolumn{2}{c|}{} & 1.0 & $55.11\pm1.03$ & $55.01\pm1.40$ & $\mathbf{61.40\pm1.20}$ & $54.82\pm0.50$ & $57.46\pm1.00$ \\ 
\hline
\multirow{6}{*}{ResNet-8} & \multirow{6}{*}{CIFAR-100} & \multirow{3}{*}{Coarse} & 0.1 & $37.73\pm0.66$ & $38.71\pm0.66$ & $35.83\pm1.90$ & $37.09\pm0.17$ & $\mathbf{53.94\pm1.10}$ \\
 &  &  & 0.5 & $54.83\pm1.02$ & $54.72\pm0.01$ & $52.55\pm0.02$ & $54.28\pm0.73$ & $\mathbf{64.46\pm0.12}$ \\
 &  &  & 1.0 & $58.07\pm1.28$ & $57.66\pm0.57$ & $55.91\pm0.04$ & $57.63\pm0.37$ & $\mathbf{66.12\pm0.19}$ \\ 
\cline{3-9}
 &  & \multicolumn{1}{c|}{\multirow{3}{*}{Fine}} & 0.1 & $40.40\pm0.94$ & $40.80\pm0.14$ & $42.54\pm0.78$ & $41.93\pm0.49$ & $\mathbf{54.27\pm0.41}$ \\
 &  & \multicolumn{1}{c|}{} & 0.5 & $47.65\pm0.31$ & $47.79\pm0.68$ & $48.76\pm0.23$ & $47.76\pm0.24$ & $\mathbf{57.06\pm0.66}$ \\
 &  & \multicolumn{1}{c|}{} & 1.0 & $47.63\pm0.24$ & $48.54\pm0.16$ & $48.75\pm0.10$ & $48.76\pm0.11$ & $\mathbf{56.47\pm0.31}$ \\
\hline
\end{tabular}
}
\end{table*}

\begin{table*}
\caption{Comparison of maximum test accuracy and communication rounds.}
\label{acc/round}
\centering
\scalebox{1.0}{
\begin{tabular}{c|l|l|c|ccccc} 
\hline
\multirow{2}{*}{Model} & \multicolumn{2}{c|}{\multirow{2}{*}{Dataset}} & \multicolumn{1}{l|}{\multirow{2}{*}{\begin{tabular}[c]{@{}l@{}}Non-IID \\ Setting ($\beta$)\end{tabular}}} & \multicolumn{5}{c}{Accuracy (\%) / Rounds} \\ 
\cline{5-9}
 & \multicolumn{2}{c|}{} & \multicolumn{1}{l|}{} & FedAvg & FedProx & SCAFFOLD & Moon & Cyclic+FedAvg \\ 
\hline

CNN-FEMNIST & \multicolumn{2}{c|}{FEMNIST} & - & 82.82 / 975  & 82.82 / 975 & 83.83 / 995  & 82.99 / 975  & $\mathbf{84.62}$ / \underline{117}  \\ 
\hline

\multirow{3}{*}{CNN-Fashion-MNIST} & \multicolumn{2}{c|}{\multirow{3}{*}{Fashion-MNIST}} & 0.1 & 81.69 / 919 & 81.52 / 919  & 82.09 / 761 & 81.67 / \underline{670}  & $\mathbf{86.19}$ / 876  \\
 & \multicolumn{2}{c|}{} & 0.5 & 86.00 / \underline{911} & 85.95 / \underline{911} & 86.25 / 980  & 86.01 / 936  & $\mathbf{88.47}$ / 954  \\
 & \multicolumn{2}{c|}{} & 1.0 & 86.58 / 929  & 86.59 / \underline{837}  & 86.53 / 961 & 86.31 / 979  &$\mathbf{88.96}$ / 860  \\ 
\hline

\multirow{3}{*}{LeNet-5} & \multicolumn{2}{c|}{\multirow{3}{*}{CIFAR-10}} & \multicolumn{1}{c|}{0.1} & 50.77 / 800 & 50.40 / 800 & \textbf{56.23} / 913 & 50.51 / \underline{553} & 54.97 / 787 \\
 & \multicolumn{2}{c|}{} & \multicolumn{1}{c|}{0.5} & 54.99 / 516 & 54.99 / 516 & \textbf{61.29} / 781 & 54.93 / 483 & 61.08 / \underline{107} \\
 & \multicolumn{2}{c|}{} & \multicolumn{1}{c|}{1.0} & 57.31 / 530 & 57.60 / 257 & \textbf{63.36} / 762 & 56.98 / 399 & 61.45 / \underline{129} \\ 
\hline
\multirow{6}{*}{ResNet-8} & \multirow{6}{*}{CIFAR-100} & \multirow{3}{*}{Coarse} & 0.1 & 44.33 / 591 & 44.96 / 591 & 43.50 / \underline{277} & 44.53 / 591 & \textbf{57.61} / 365 \\
 &  &  & 0.5 & 56.57 / 934 & 56.47 / 816 & 56.15 / 459 & 56.95 / 955 & \textbf{66.59} / \underline{380} \\
 &  &  & 1.0 & 58.39 / 977 & 58.18 / 836 & 58.96 / 624 & 58.01 / 858 & \textbf{67.37} / \underline{265} \\ 
\cline{3-9}
 &  & \multicolumn{1}{c|}{\multirow{3}{*}{Fine}} & 0.1 & 42.83 / 699 & 42.66 / 836 & 46.95 / 360 & 43.17 / 755 & \textbf{56.11} / \underline{252} \\
 &  & \multicolumn{1}{c|}{} & 0.5 & 48.86 / 394 & 48.85 / 683 & 51.48 / 323 & 49.21 / 362 & \textbf{58.92} / \underline{150} \\
 &  & \multicolumn{1}{c|}{} & 1.0 & 49.88 / 489 & 50.00 / 607 & 51.64 / 548 & 50.18 / 493 & \textbf{58.33} / \underline{128} \\
\hline
\end{tabular}
}
\vspace{-0.1in}
\end{table*}

\section{Performance Evaluation} \label{sec:exp}

We compared our cyclic pre-training method with the random initialization method on four representative FL algorithms. The datasets we used include four image benchmarks. Our experiments were conducted on a Ubuntu workstation with an Intel i9-12900K CPU, and an NVIDIA GeForce RTX 3090Ti GPU.
To show the advantages of the cyclic pre-training approach, we designed experiments to answer the following four research questions: 
\\
\textbf{RQ1 (Classification performance)}: What are the advantages of cyclic pre-training on experiments?
\\
\textbf{RQ2 (Convergence performance)}: How does the pre-trained model affect convergence?
\\
\textbf{RQ3 (Impact on loss landscapes)}: How does cyclic pre-training affect the final loss landscape?

\subsection{Experimental Settings}
\noindent \textbf{Datasets.}
We evaluated the performances of our CyclicFL on four image classification tasks, i.e., FEMNIST \cite{Caldas-et-al:leaf},
Fashion-MNIST \cite{Xiao-et-al:fashmnist}, 
CIFAR-10, CIFAR-100 \cite{Krizhevsky-et-al:cifar}. 
Note that each image in CIFAR-100 is tagged with ``Coarse Label'' and ``Fine Label''. We use ``C'' and ``F'' to represent ``Coarse'' and ``Fine'', respectively, to fully exploit the data.
For heterogeneity configurations, we considered Dirichlet distribution throughout this paper, which is parameterized by a coefficient $\beta$, denoted as $Dir(\beta)$. $\beta$ determines the degree of data heterogeneity. The smaller the $\beta$ is, the more heterogeneous the data distributions will be. We created three Non-IID scenarios by setting $\beta$ as $0.1$, $0.5$, and $1.0$, respectively. Note that FEMNIST is generated with a Non-IID setting.

\noindent \textbf{Models.}
To show the scalability of cyclic pre-training to various types of models, we utilized the LeNet-5 \cite{LeCun-et-al:Lenet}, which is composed of two convolutional layers followed by three full-connected layers, for the CIFAR-10 dataset. For CIFAR-100, we used ResNet-8 \cite{He-et-al:he2016deep} with batch normalization.
For FEMNIST, we applied a model consisting of two convolutional layers and one fully-connected layer. A model composed of two convolutional layers followed by a dropout layer and two fully-connected layers are used for Fashion-MNIST.

\noindent \textbf{Baselines.}
Since cyclic pre-training is a pre-training method, we conducted it to obtain a pre-trained model, then utilized FedAvg \cite{McMahan-et-al:fedavg} as the default FL algorithm to continue FL training. 
We denoted the default algorithm as ``Cyclic+FedAvg''. 
Our method is evaluated on four representative FL algorithms: FedAvg \cite{McMahan-et-al:fedavg}, FedProx \cite{Li-et-al:fedprox}, SCAFFOLD \cite{Karimireddy-et-al:scaffold}, and Moon \cite{Li-et-al:moon}.

\noindent \textbf{Hyperparameters.}
\label{sgdhyper}
Since our method is accomplished before conventional FL algorithms, we elaborated individual hyperparameters for P1 and P2, respectively. In both phases, we set the total number of devices to 100, except for FEMNIST. FEMNIST is naturally attributed to 190 devices.
In P1, we uniformly selected 25\% of the total devices to participate in local training in every round by default. The total training rounds in P1 are set to 100 ($T_{c}$). The maximum local update steps are set as 20.
In P2, the ratio of devices participating in local training in every round is set as 10\%.
Note that FedProx and Moon both introduce extra hyperparameters that need to be tuned. 
We set the coefficient $\mu$ in FedProx as 0.01. For Moon, we set the distillation temperature $T$ as 0.5 and $\mu$ as 0.1.
For tasks except for CIFAR-100, we configured the learning rate as 0.01, momentum as 0, and weight decay as 0. For CIFAR-100, the learning rate is 0.1, the momentum is 0.5, and the weight decay is configured as 1e-3. The learning rate decays to 99.8\% after each round. 
For local training, we set 5 as the local epochs. The training rounds are 1000 ($T_{total}$) for all experiments. For both P1 and P2, we set the batch size to 32.
For the ablation study, the experiments are conducted on CIFAR-10 and the algorithm is Cyclic+FedAvg. We use $\beta=0.5$ for the ``impacts of the pre-training rounds" and $\beta=0.1$ for the rest.

\subsection{Experimental Results}\label{sec5}

\noindent \textbf{Classification performances.}
\label{fastconv}
Table~\ref{exp_randomseeed} shows the top-1 test accuracy and variances for all FL baselines under Non-IID settings.
We can observe that Cyclic+FedAvg greatly beat four basic FL algorithms for the CIFAR-100-Coarse/Fine dataset. For CIFAR-100-Coarse with $\beta=0.5$, Cyclic+FedAvg outperforms FedAvg, FedProx, SCAFFOLD, and Moon by $9.63\%$, $9.74\%$, $11.91\%$, and $10.18\%$, respectively.
For the Fashion-MNIST dataset, the cyclic pre-training also consistently outperforms all baselines.
For CIFAR-10 and FEMNIST, we observe that SCAFFOLD consistently outperforms other methods, including Cyclic+FedAvg. Since our method is a pre-training process, the higher classification accuracy of SCAFFOLD compared to Cyclic+FedAvg is not against the effectiveness of cyclic pre-training.

Table~\ref{acc/round} presents the maximum accuracy and the number of training rounds consumed to achieve it for all baselines. The number of training rounds is underlined for better visualization. We find that cyclic+fedAvg achieves the highest accuracy in most cases while maintaining the smallest communication rounds consumed.
For experiments including CIFAR-10 with $\beta=0.5$, CIFAR-10 with $\beta=1.0$, CIFAR-100-Fine with $\beta=0.5$, and CIFAR-100-Fine with $\beta=1.0$, we can observe a tremendously fast convergence speed when starting training from a cyclic pre-training global model: $7$ more rounds are consumed to achieve an accuracy of $61.08\%$; $29$ more rounds to achieve an accuracy of $61.45\%$; $50$ more rounds to achieve $58.92\%$; and $28$ more rounds to achieve $58.33\%$. 
For the FEMNIST dataset, we can find that it only consumes $17$ more rounds to achieve an accuracy of $84.62\%$. 

\begin{figure*}[h]
\centering
\subfigure[$\beta=0.1$]{\includegraphics[width=.32\linewidth]{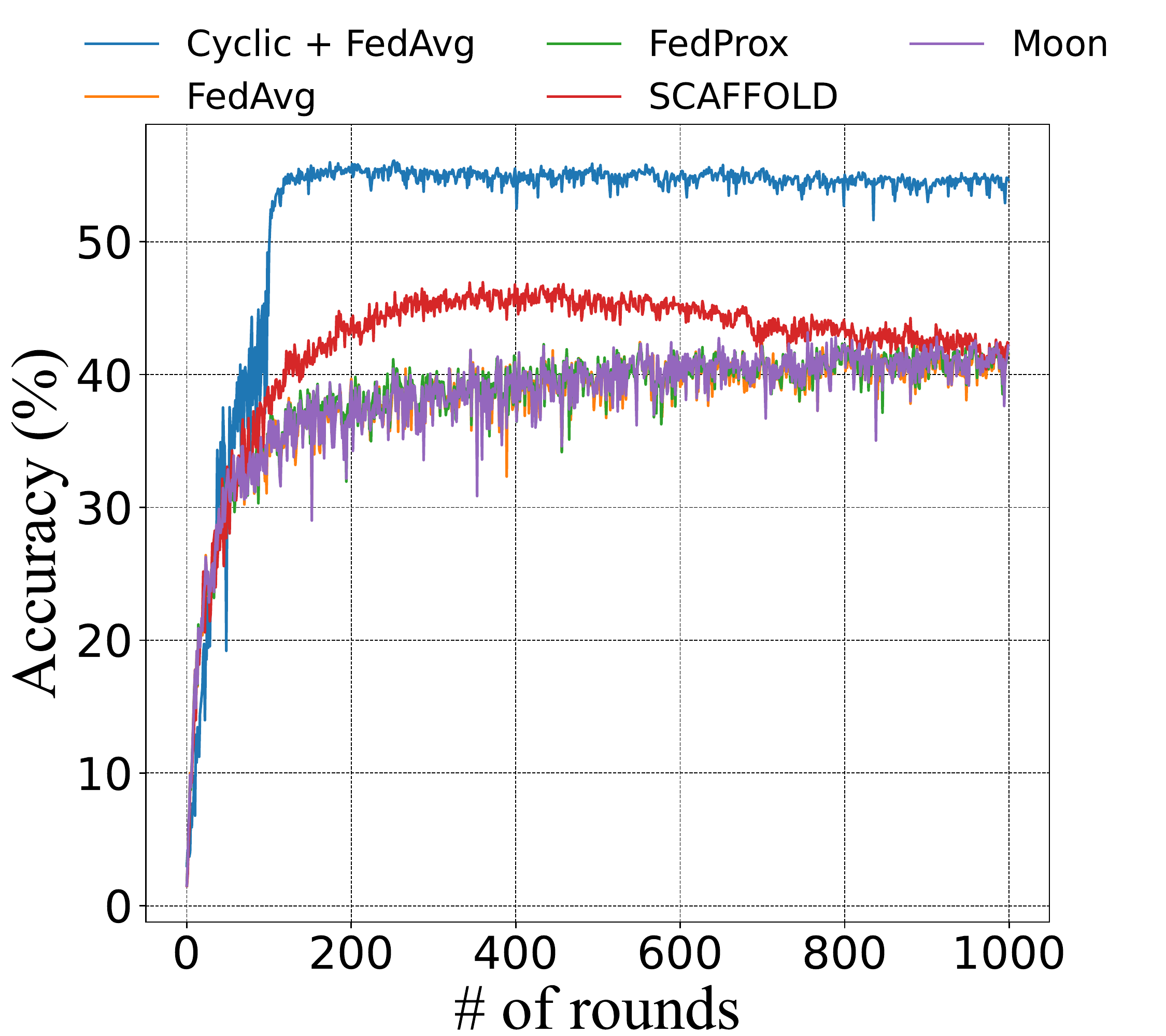}}
\subfigure[$\beta=0.5$]{\includegraphics[width=.32\linewidth]{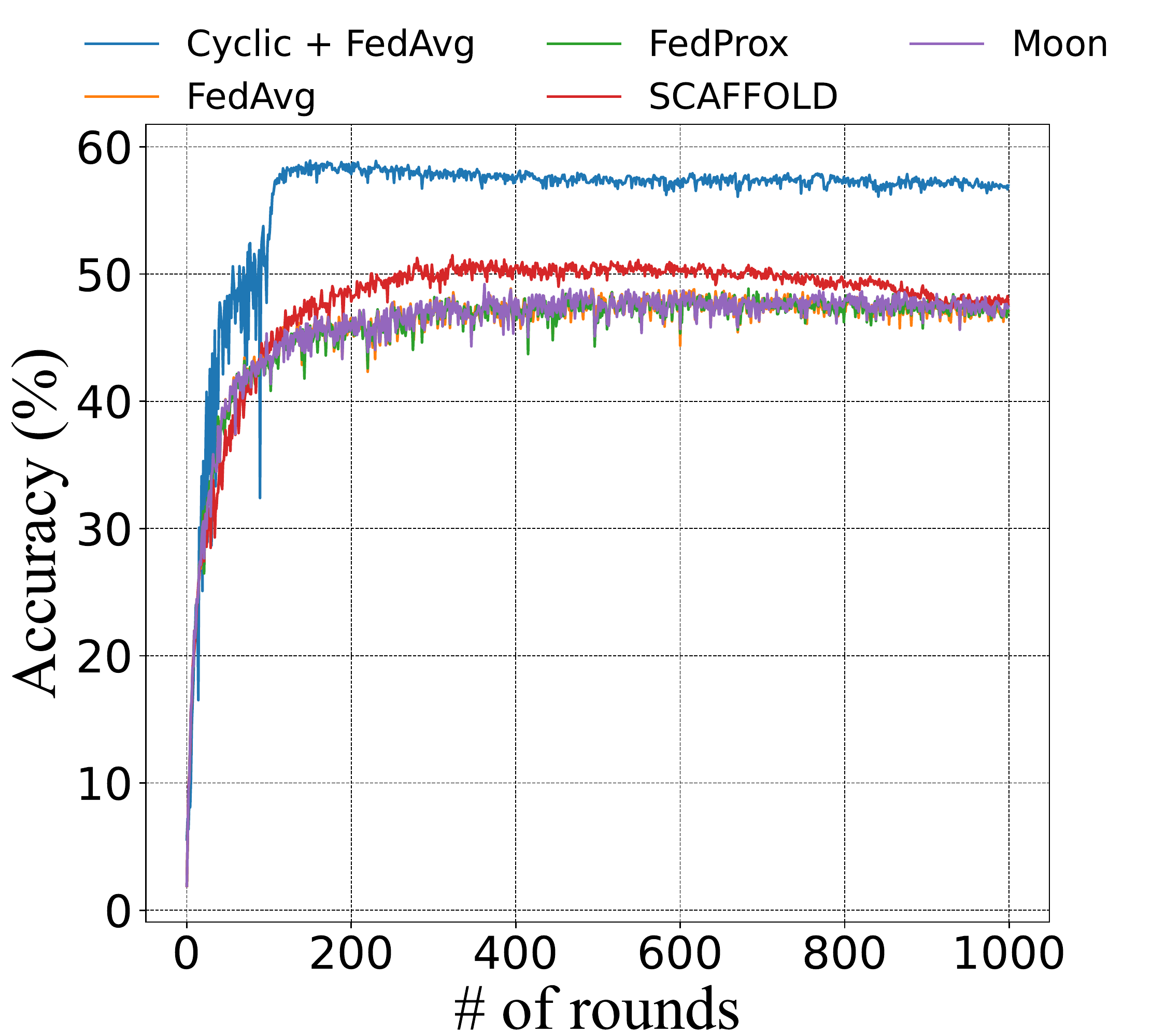}} 
\subfigure[$\beta=1.0$]{\includegraphics[width=.32\linewidth]{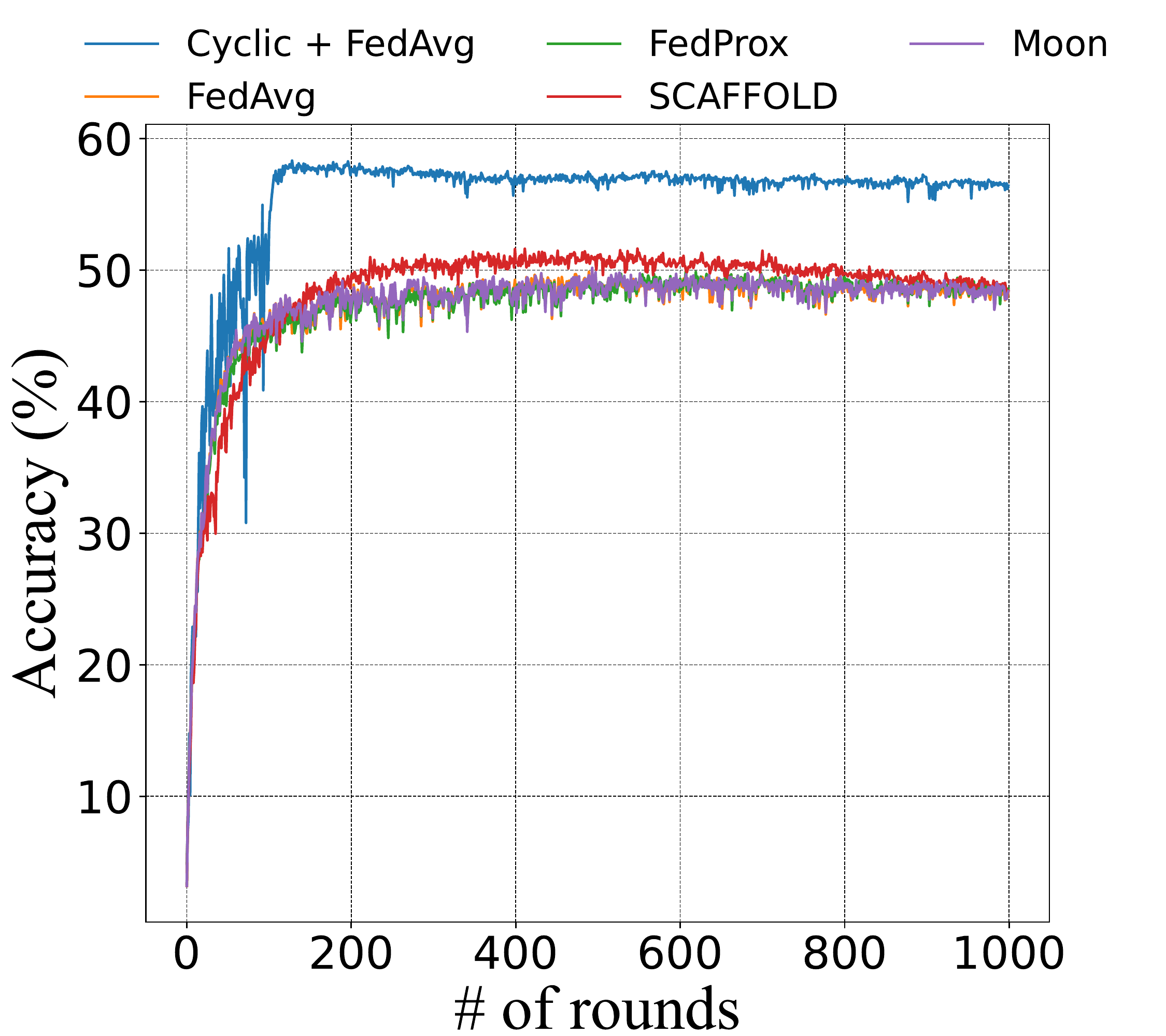}}  \\
\subfigure[$\beta=0.1$]{\includegraphics[width=.32\linewidth]{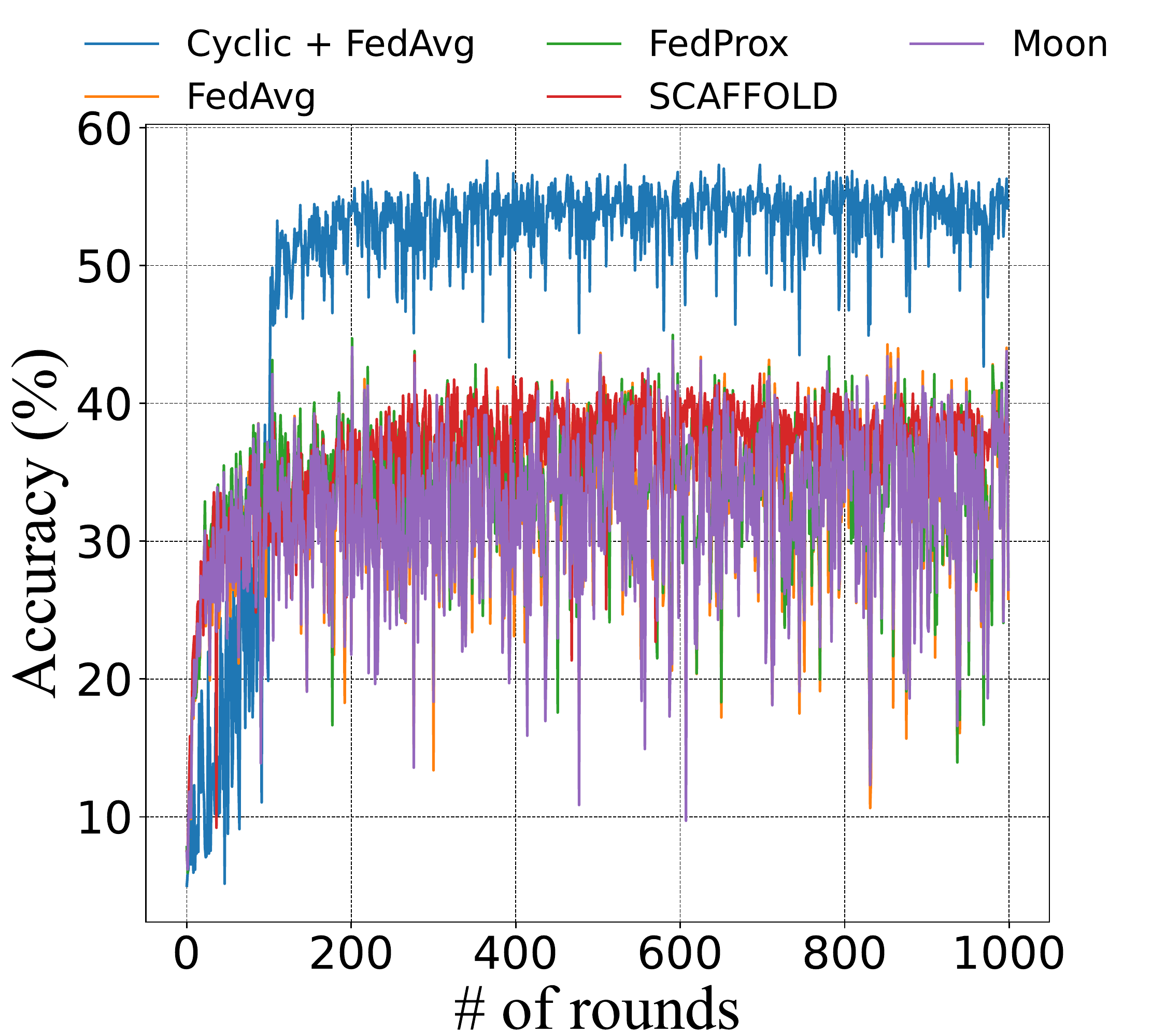}} 
\subfigure[$\beta=0.5$]{\includegraphics[width=.32\linewidth]{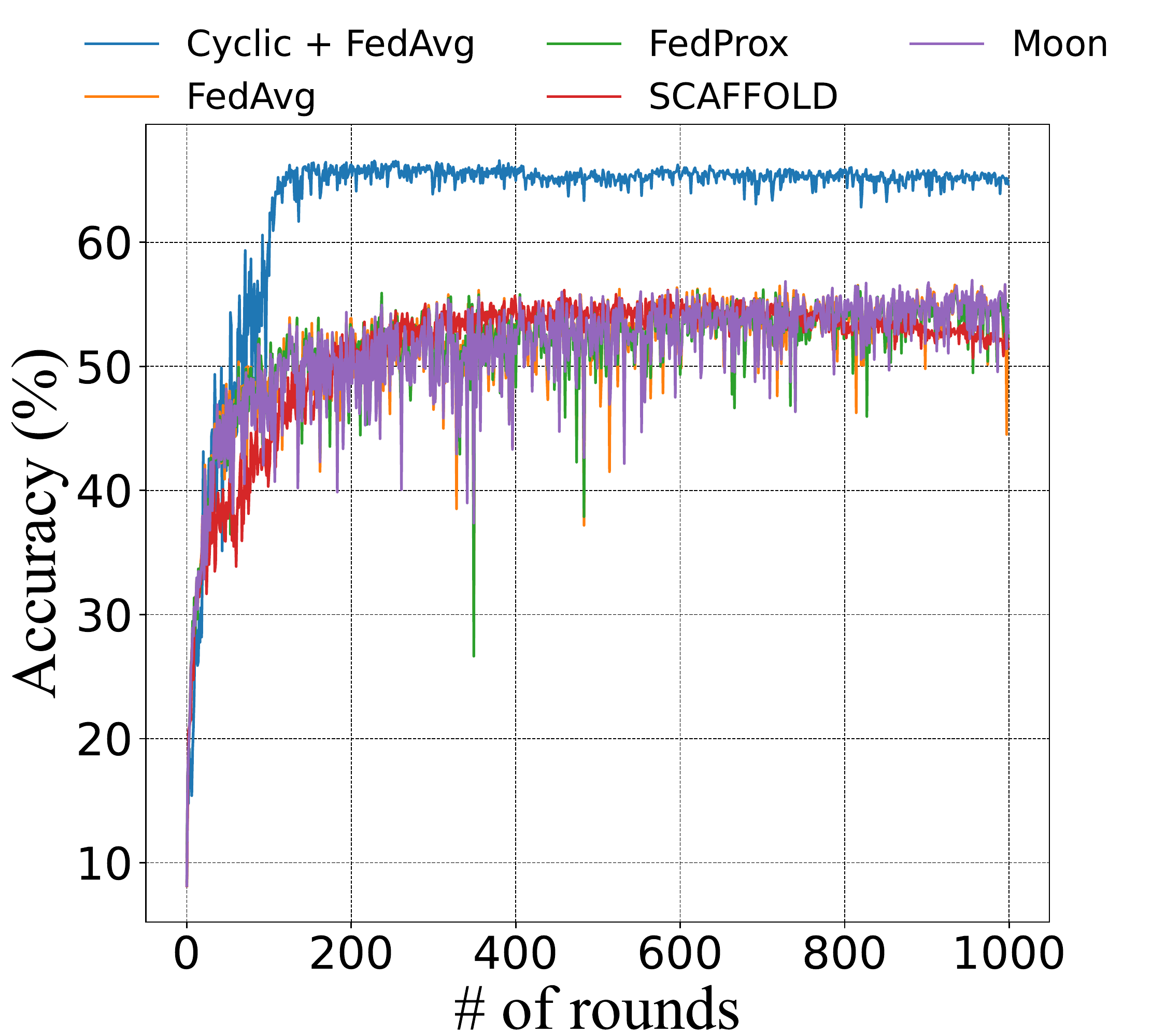}}
\subfigure[$\beta=1.0$]{\includegraphics[width=.32\linewidth]{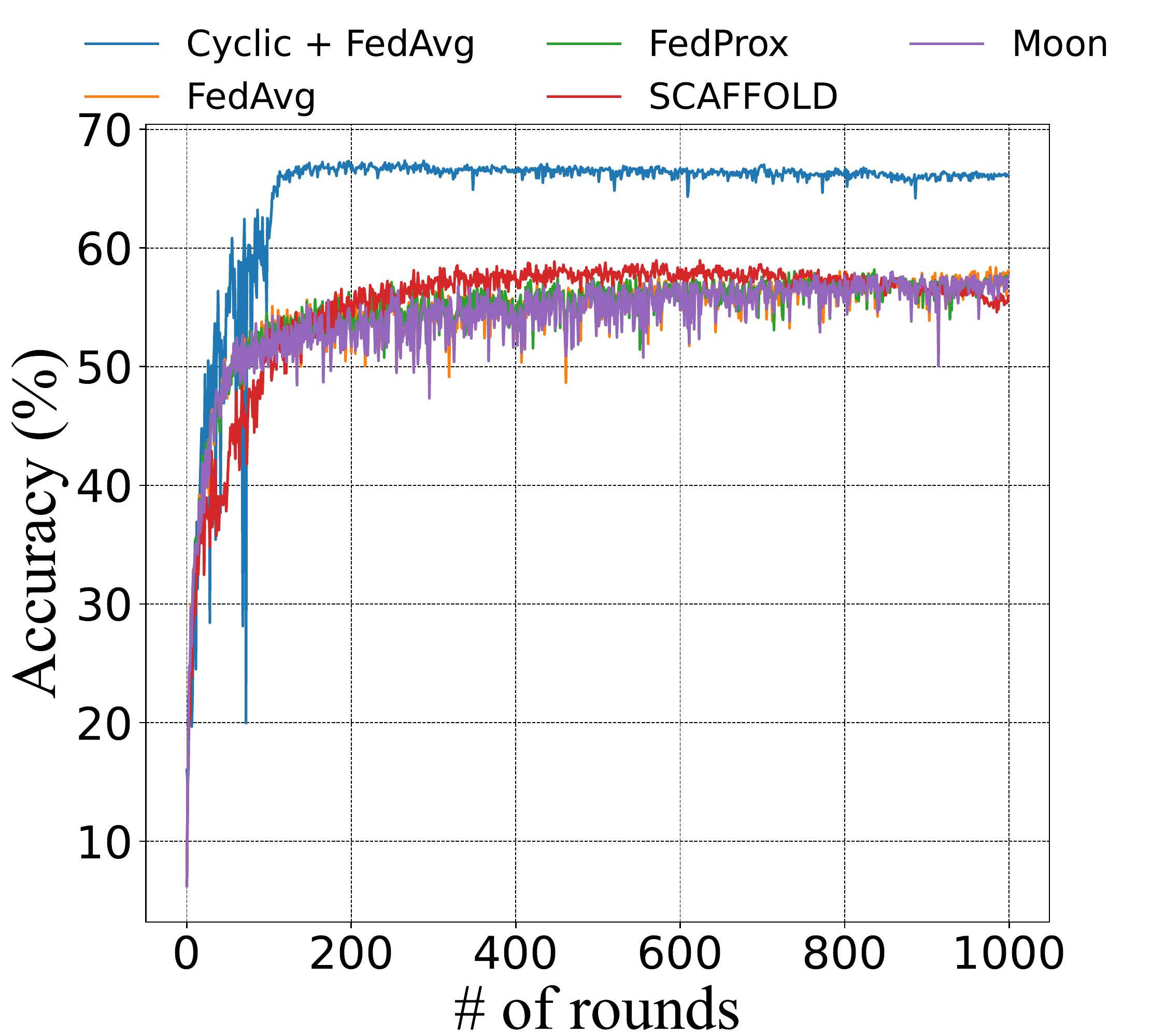}}
\vspace{-0.1in}
\caption{Test accuracy for CIFAR-100-Fine. 
}
\label{accuracy}
\end{figure*}

\begin{figure}[h]
    \centering
    \includegraphics[width=0.8\linewidth]{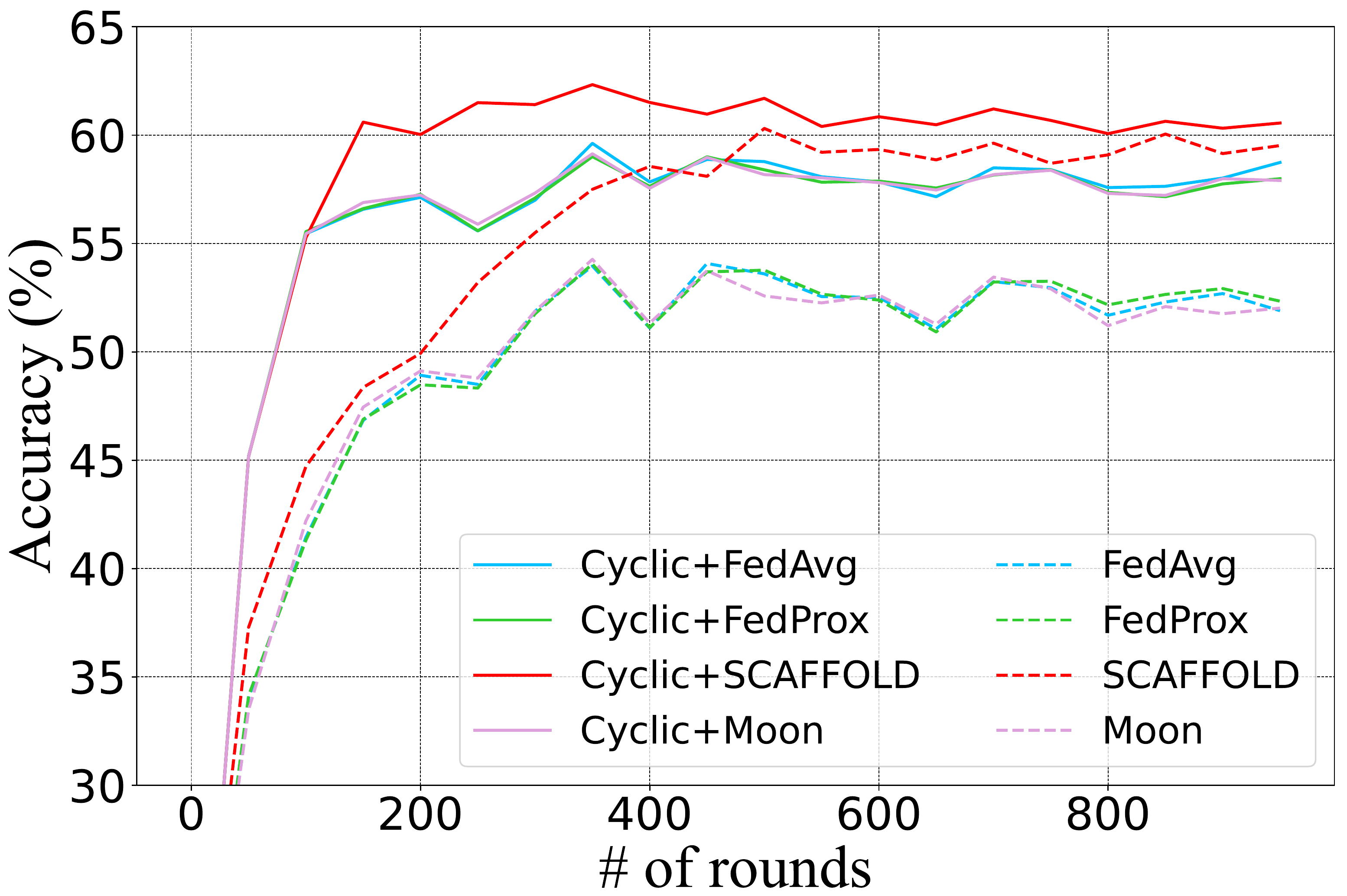}
    \vspace{-0.1in}
    \caption{Comparison of test accuracy for CIFAR-10, $\beta=0.5$.}
   \vspace{-0.25in}
    \label{acc improve}
\end{figure}

In Figure~\ref{accuracy}, we show the learning curves of experiments on the CIFAR-100-Fine. 
The first $100$ rounds represent the cyclic pre-training process, and we switched to P2 to complete the rest of the training process. 
In this figure, we can observe a notable accuracy leap that occurs after switching to P2, which highlights the effectiveness of cyclic pre-training in significantly accelerating convergence speed.
Table~\ref{cyclic+} compares the accuracy and consumed communication rounds of Cyclic+Y, where Y represents FL baselines (i.e., FedAvg, FedProx, SCAFFOLD, and Moon).
Figure~\ref{acc improve} presents the corresponding learning curves. For better visualization, we show the data points every 50 rounds.
The four FL algorithms are depicted with dash-dotted lines, and their Cyclic+ versions are plotted in solid lines.
Both Table~\ref{cyclic+} and Figure~\ref{acc improve} show that cyclic pre-training is compatible with these representative FL algorithms, and it improves classification accuracy on FedProx, Moon, and SCAFFOLD by $4.84\%$, $5.62\%$, and $0.33\%$, respectively. 


\begin{table}[h] 
\vspace{-0.05in}
\centering
\caption{Comparison of maximum test accuracy and consumed communication rounds.}
\label{cyclic+}
\scalebox{1.0}{
\begin{tabular}{c|c|l} 
\hline
\multirow{2}{*}{Algorithm} & \multicolumn{2}{c}{Accuracy (\%) / Rounds} \\ 
\cline{2-3}
 & w/o Cyclic+ & \multicolumn{1}{c}{w/ Cyclic+} \\ 
\hline
FedAvg &54.99 / 466  &61.08\textbf{(+6.09)} / 107\textbf{(-359)} \\
FedProx &54.99 / 516  &61.09\textbf{(+6.10)} / 107\textbf{(-409)}  \\
Moon &54.93 / 483  &61.17\textbf{(+6.24)} / 107\textbf{(-376)} \\
SCAFFOLD &61.29 / 781 &62.68\textbf{(+1.39)} / 232\textbf{(-549)} \\
\hline
\end{tabular}}
 \vspace{-0.05in}
\end{table}

\begin{figure}[h]
 \vspace{-0.1in}
\centering
\subfigure[Initial-Randomly]{\includegraphics[width=.49\linewidth]{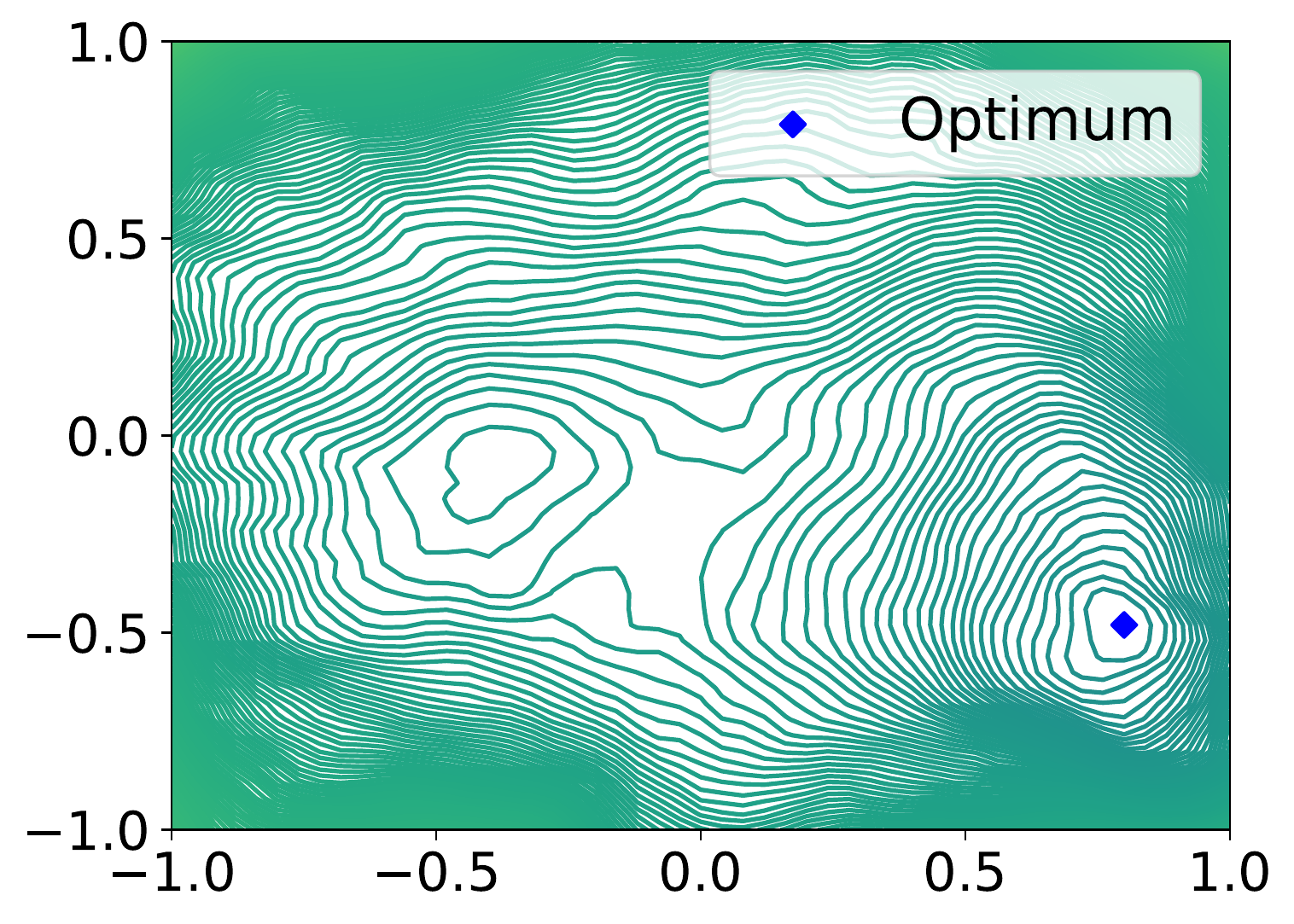}}
\subfigure[Final-Randomly]{\includegraphics[width=.49\linewidth]{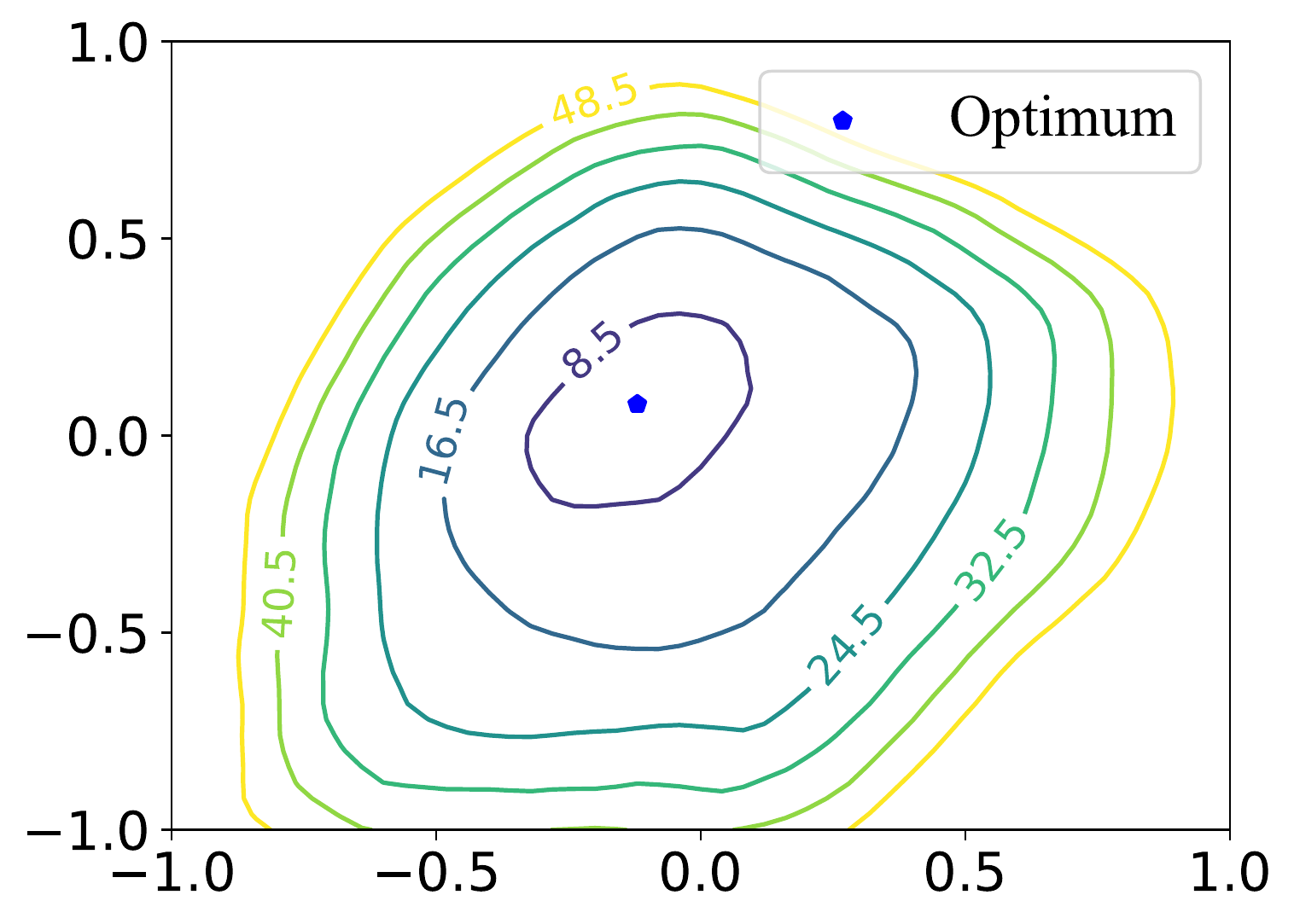}} \\
\subfigure[Initial-Pretrained]{\includegraphics[width=.49\linewidth]{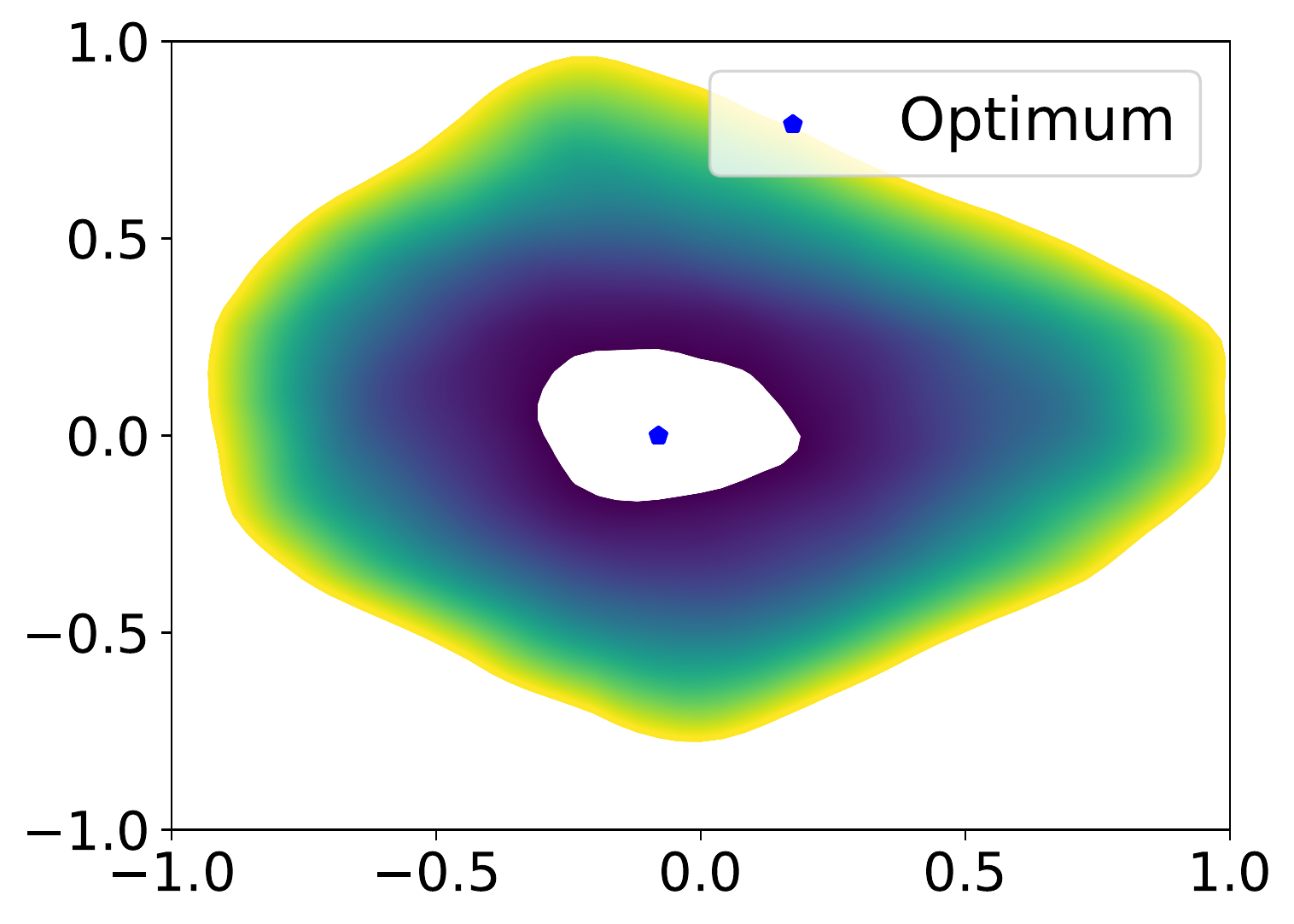}}
\subfigure[Final-Pretrained]{\includegraphics[width=.49\linewidth]{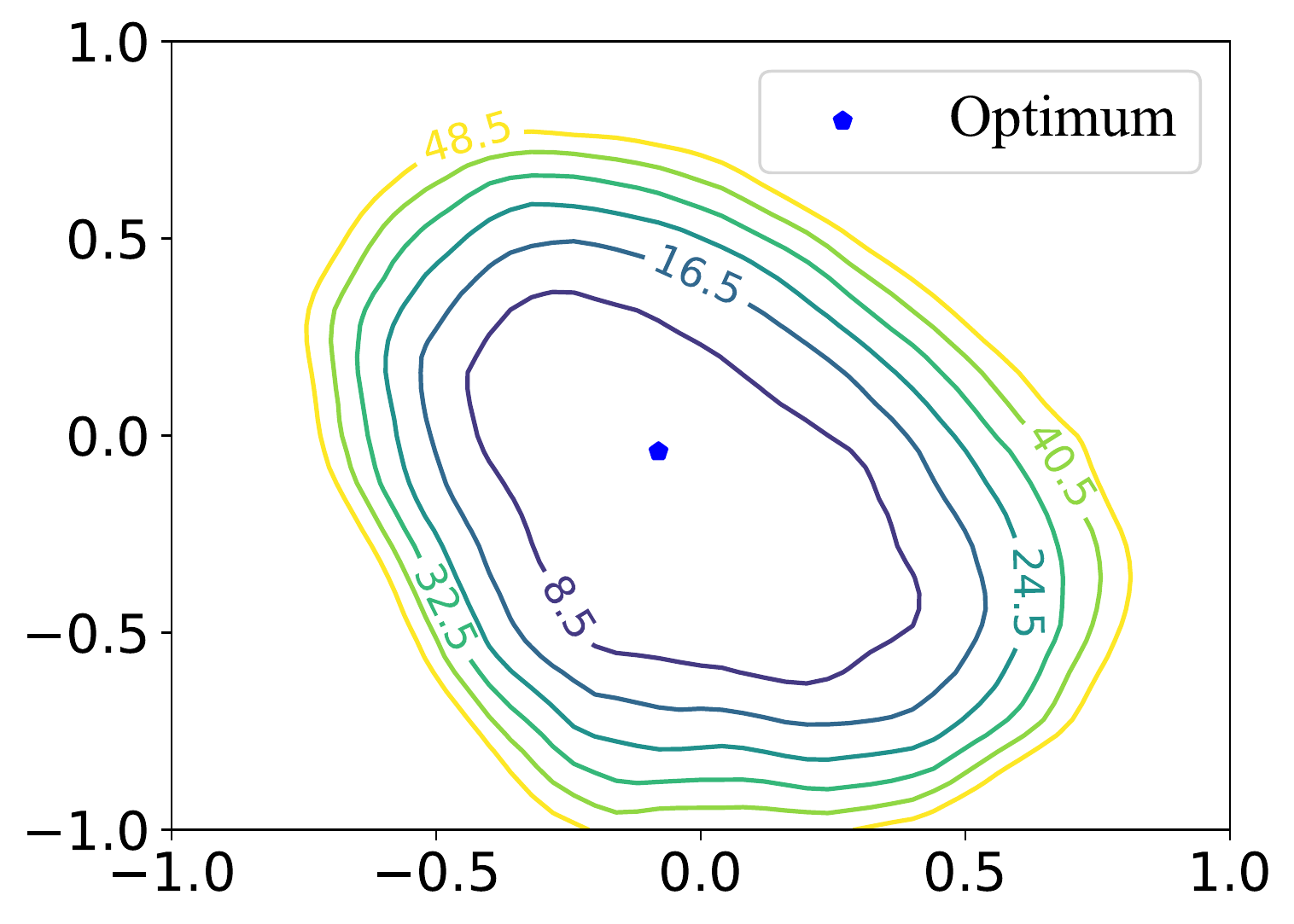}}
\vspace{-0.1in}
\caption{Loss landscape visualizations of a LeNet-5 model for CIFAR-10. $\beta$ is set to $0.5$. 
}
\label{lossland}
\vspace{-0.25in}
\end{figure}

\noindent \textbf{Loss landscapes of global models.} Figure~\ref{lossland} visualizes the loss landscapes of models. 
We conducted experiments on CIFAR-10 using random and pre-trained weights from cyclic pre-training. We projected the trained global models onto their loss landscapes based on the neural network visualization method \cite{Li-et-al:lossland}. The ($0.0$, $0.0$) coordinate represents the loss value given the current state of weights.
Subfigure.~\ref{lossland} (a) describes the loss landscapes of the global model initialized with random weights, and subfigure.~\ref{lossland} (c) shows that initialized with pre-trained weights. Note that the scales of the contour of the two subfigures are equivalent. Subfigures.~\ref{lossland} (b) and ~\ref{lossland} (d) show the loss landscapes of their final global models, respectively.
When comparing the randomly initialized model to the one created through cyclic pre-training, we can see that the latter has flatter loss basins and lower loss values at the beginning of training.
From the perspective of the loss landscapes at convergence, global models without cyclic pre-training converge to sharp loss basins with larger loss values.

\noindent\textbf{Communication efficiency}.
Table~\ref{comm} shows the comparison of communication consumptions between different FL algorithms.
We use notations $\mathbf{X}$ and $T_{res}=T_{total}-T_{c}$ to represent the capacity of a model and the training rounds consumed in P2, respectively.
In P1, the communication overhead is controlled by model transmitting between the server and devices. Two transmissions are required in each round, leading to a communication consumption of 2$S_c\mathbf{X}$. Therefore, the overall overhead is 2$S_cT_{c}\mathbf{X}$ in P1. FedAvg, FedProx, and Moon all consume 2$S_pT_{c}\mathbf{X}$. SCAFFOLD demands uploading control variables, thus leading to an overhead of 4 $S_pT_{c}\mathbf{X}$ in P1. 
Since a small value of $T_c$ in P1 can already provide significant gains, the communication overhead is dominated by the federated training. 
We further tested the transmission speed between two devices. Note that we only conduct the pre-training in P1. 
Take the LeNet and ResNet-8 as examples.
The average wall clock transmission times of the two models in P1 are 190ms and 236ms, respectively. Since the average wall clock training time per round for P1 and P2 are 1425ms and 3122ms, respectively, the transmission expense in P1 is negligible. Overall, our approach will not be dragged by the sequential model transmission process.

\begin{table}[h] 
\centering
\caption{Comparison of communication overhead.}
\label{comm}
\scalebox{1.0}{
\begin{tabular}{c|c|l} 
\hline
\multirow{2}{*}{Algorithms} & \multicolumn{2}{c}{Communication Overhead} \\ 
\cline{2-3}
 & w/o Cyclic+ & \multicolumn{1}{c}{w/ Cyclic+} \\ 
\hline\hline 
FedAvg &$2$$S_pT_{total}\mathbf{X}$  &$2$$\left[S_cT_{c}+S_pT_{res}\right]$$\mathbf{X}$ \\
FedProx &$2$$S_pT_{total}\mathbf{X}$ &$2$$\left[S_cT_{c}+S_pT_{res}\right]$$\mathbf{X}$ \\
Moon &$2$$S_pT_{total}\mathbf{X}$  &$2$$\left[S_cT_{c}+S_pT_{res}\right]$$\mathbf{X}$ \\
\multicolumn{1}{c|}{SCAFFOLD} & \multicolumn{1}{c|}{$4$$S_pT_{total}\mathbf{X}$} & \multicolumn{1}{c}{$2$$\left[S_cT_{c}+2S_pT_{res}\right]$$\mathbf{X}$} \\
\hline
\end{tabular}}
\end{table}
\vspace{-0.25in}

\subsection{Ablation Study}

\begin{figure}[htbp]
    \centering
    \includegraphics[width=0.8\linewidth]{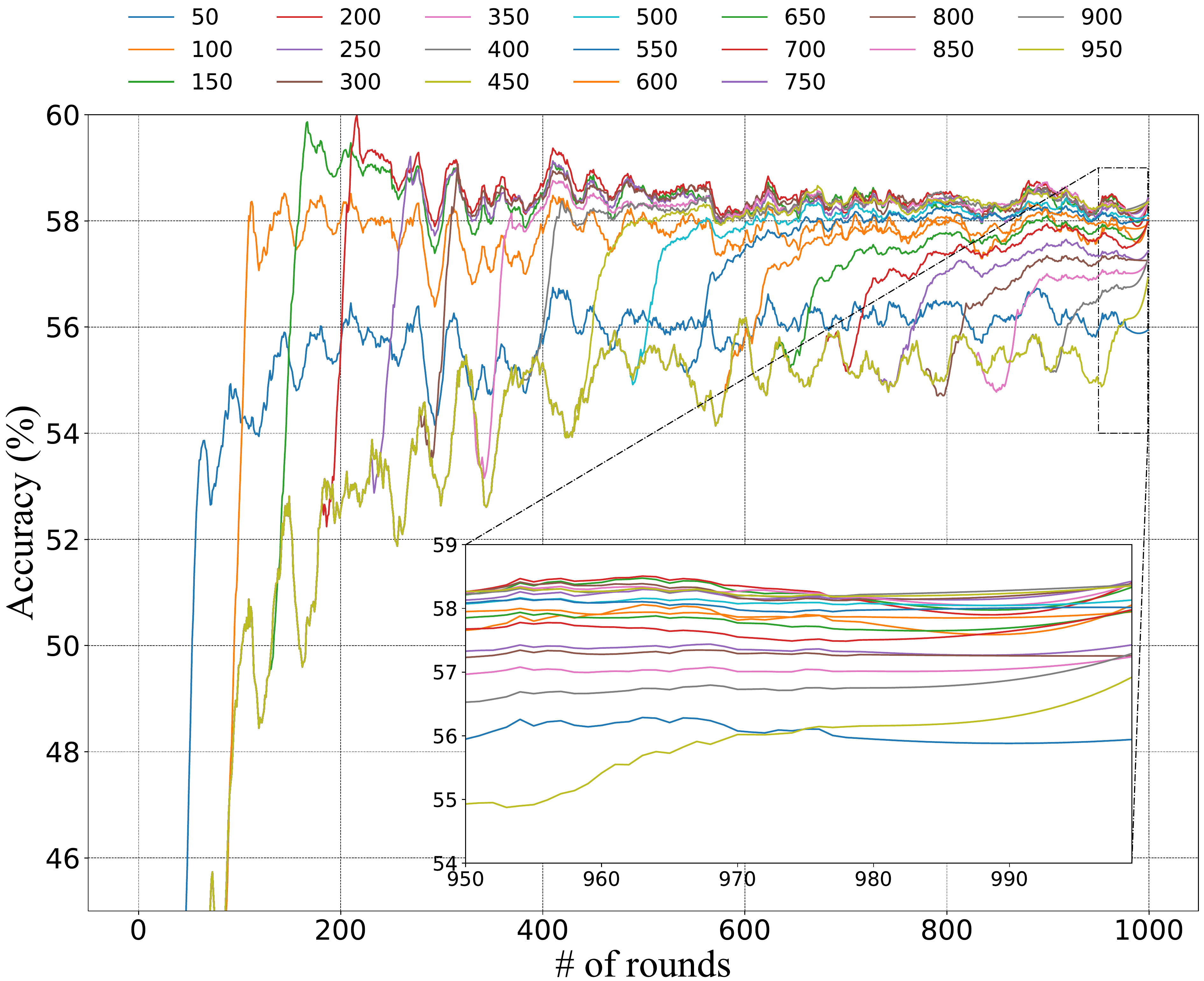}
    \caption{Learning curves when switching from P1 to P2 at different rounds. The final 50 rounds are zoomed in.}
    \label{allcurve}
\end{figure}

\begin{figure}[htbp]
\centering
\includegraphics[width=0.75\linewidth]{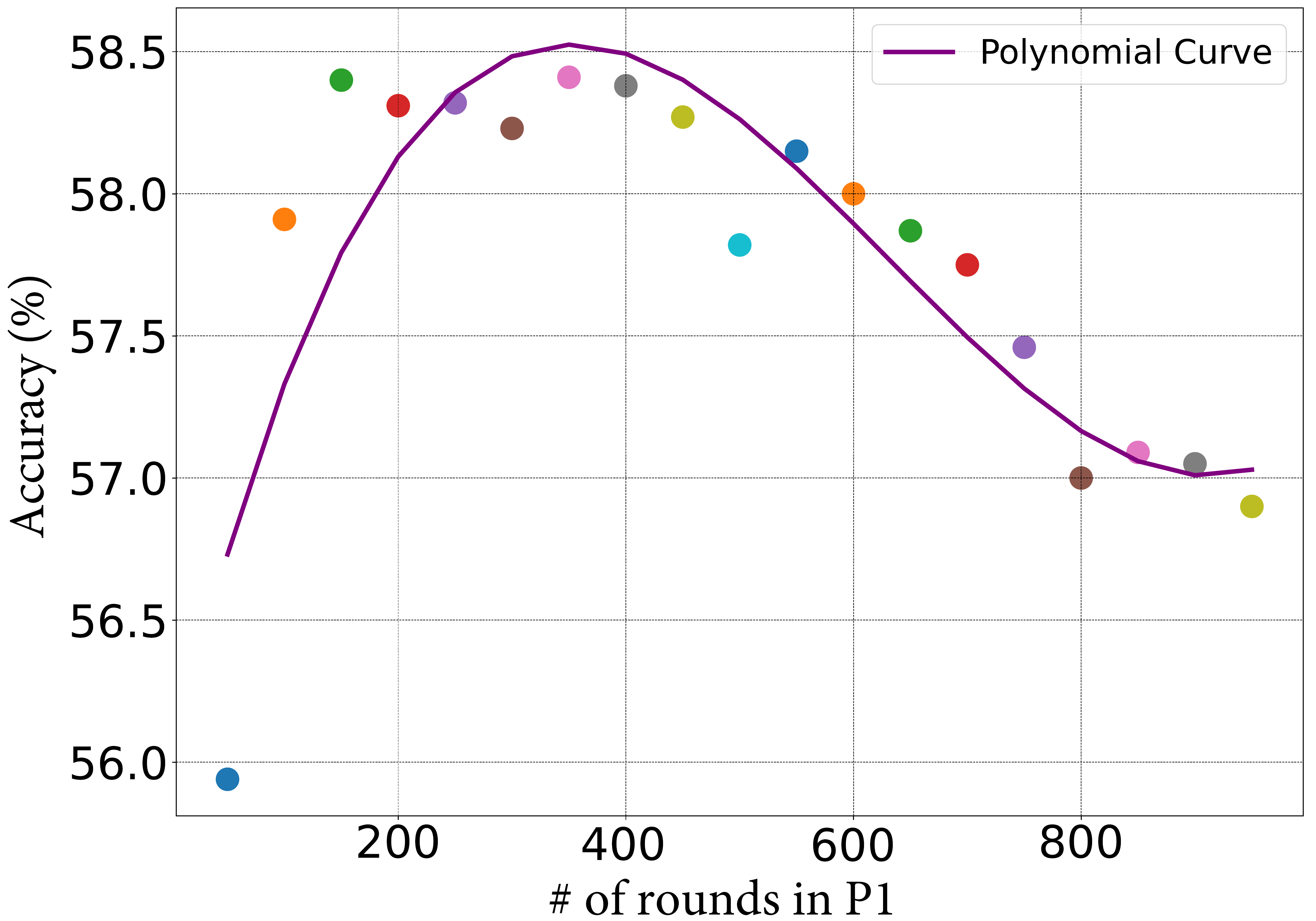}
\caption{Final test accuracy of training based on different training rounds in P1.}
\label{finalfitting}
\vspace{-0.25in}
\end{figure}

\noindent \textbf{Impacts of the pre-training rounds.}
Figure~\ref{allcurve} compares final test accuracy based on different training rounds spent in P1 while 
maintaining the maximum local training steps unchanged. The total rounds (P1+P2) are $1000$.
To better visualize the trends of learning curves, all curves are smoothed by the Savitzky-Golay filter~\cite{Savitzky-et-al:filter}. The window size of the filter is $41$, and the order is $3$. We can observe that the fast convergence phenomenon occurs in all scenarios.
Although cyclic pre-training already brings accuracy improvements compared to vanilla FL (e.g., extend cyclic pre-training up to $950$ rounds), we can still obtain more improvements in accuracy if we initiate a switching point (e.g., switch to vanilla FL at the $950^{th}$ rounds).
Figure~\ref{finalfitting} shows the relation between the number of training rounds in P1 and the final test accuracy by scatters.
We fit the scatters with a $3^{rd}$ order polynomial curve, which shows the trend of the final accuracy when switching from P1 to P2 at different rounds. 
Similar to the phenomenon in \cite{Liu-et-al:transferunder}, we can find that the transferability of the pre-trained model rapidly increases at early rounds, followed by a slow descent.
Although increasing P1 from $100$ rounds to $150$ rounds can improve accuracy, the advantage of convergence acceleration is diminished due to the extra rounds consumed in P1. Therefore, a trade-off between efficiency and accuracy performance needs to be considered in practical situations. 

\begin{figure}[h]
\vspace{-0.1in}
\centering
\subfigure[Dataset distributions]{\includegraphics[width=.48\linewidth]{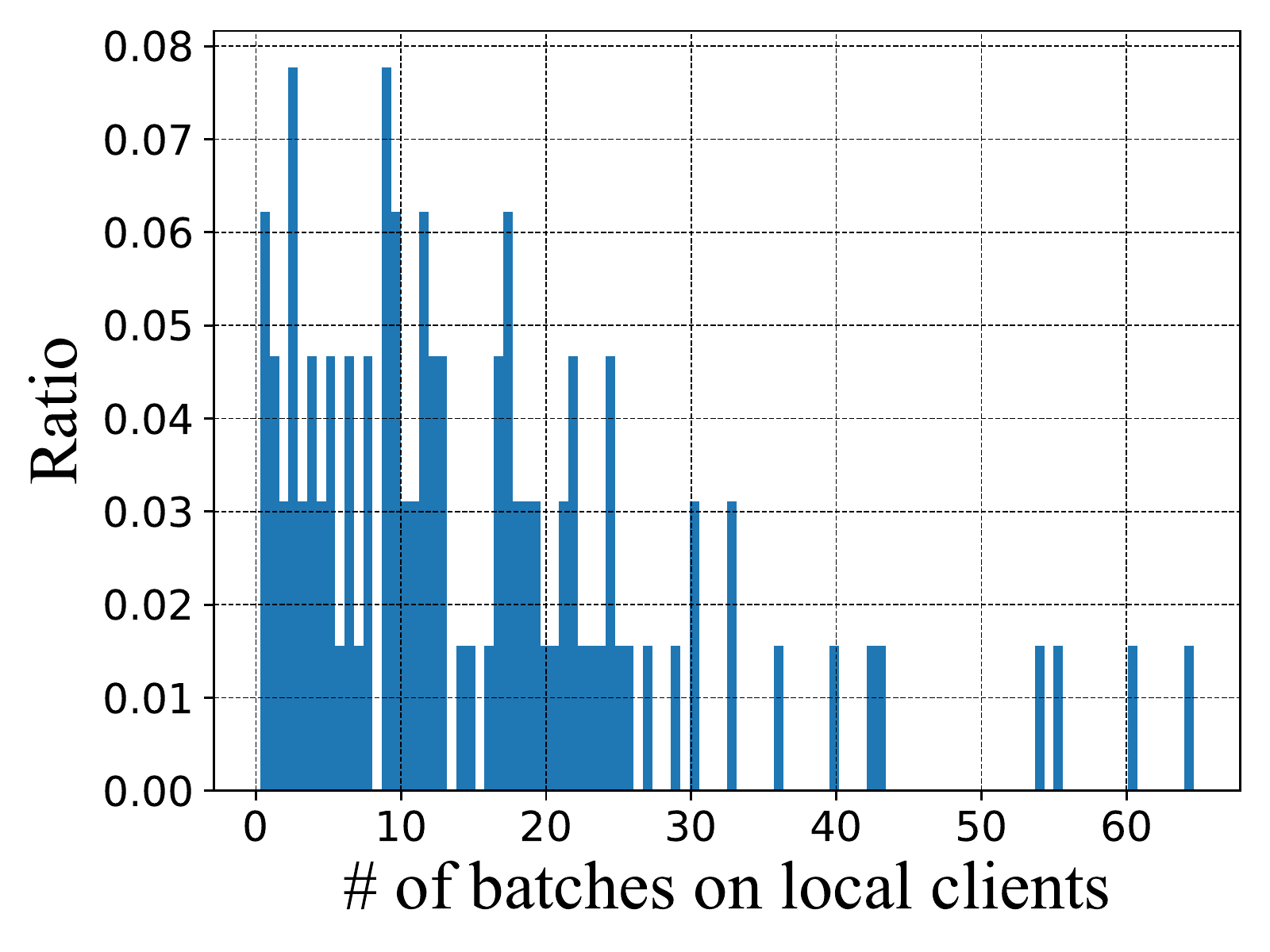}
}
\subfigure[Accuracy v.s. local steps]{\includegraphics[width=.48\linewidth]{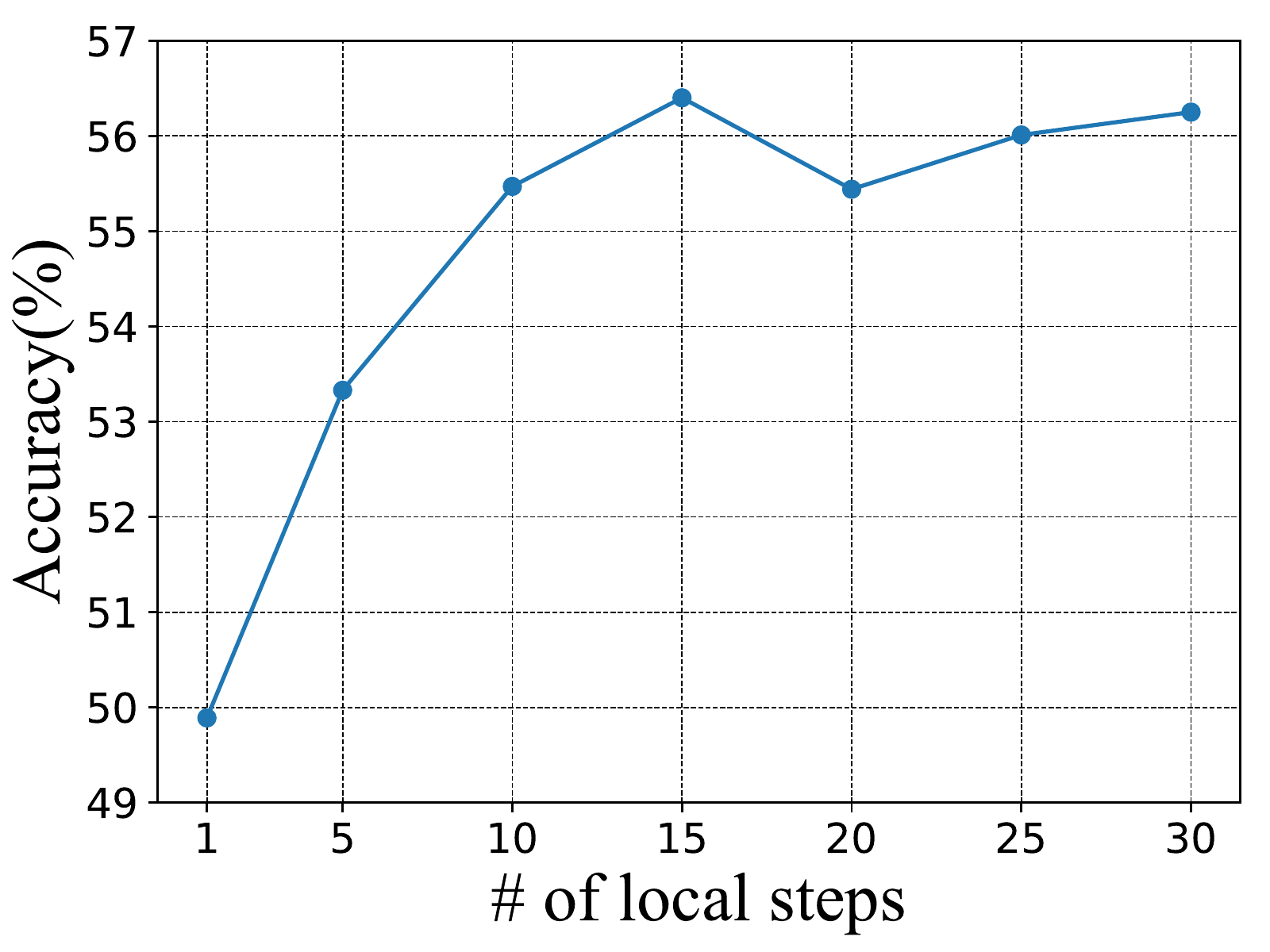}} 
\vspace{-0.1in}
\caption{(a) Distributions of the \# of batches. (b) Maximum accuracy based on different \# of local steps in P1.}
\label{dis_local_steps}
\vspace{-0.1in}
\end{figure}

\noindent \textbf{Impacts of the pre-training local steps.} Subfigure (a) in Figure.~\ref{dis_local_steps} shows the distribution of the number of local batches. 
Subfigure (b) shows the maximum accuracy under different values of local training steps in P1. We can observe that the maximum accuracy is significantly improved as we increase the local steps from $1$ to $10$. Nevertheless, increasing steps from $10$ to $30$ exhibits marginal accuracy improvements. 
This phenomenon aligns with the distribution of the number of device-owned batches since most devices have less than $30$ batches. Increasing the local steps will not bring a notable boost to the coverage of whole local data.



\begin{figure}[h]
\vspace{-0.1in}
\centering
\includegraphics[width=1.0\linewidth]{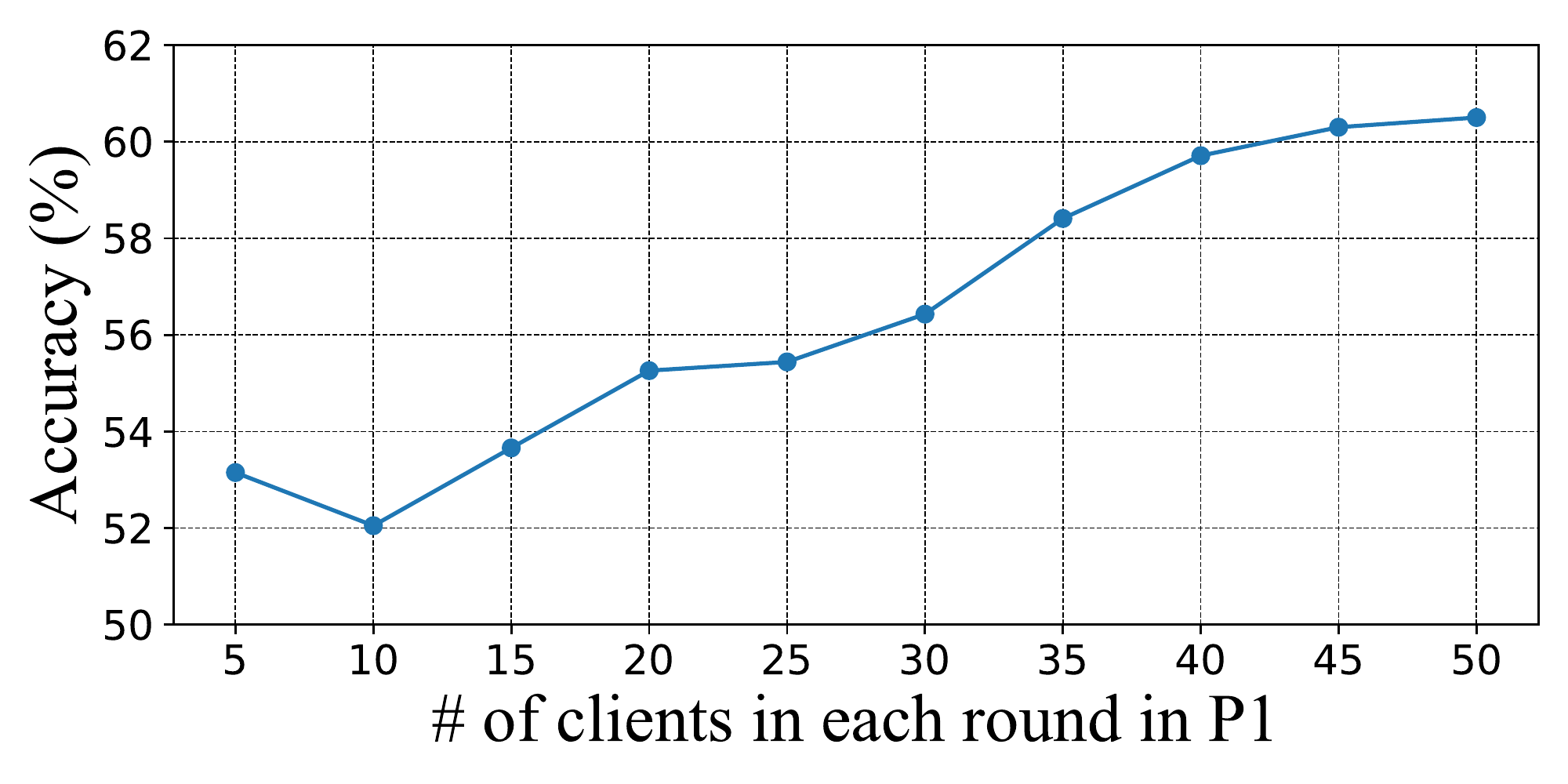}
\vspace{-0.25in}
\caption{Maximum accuracy based on different \# of devices in each round in P1.}
\label{n_device_p1}
\vspace{-0.1in}
\end{figure}

\noindent \textbf{Impacts of the number of devices in P1.}
Figure.~\ref{n_device_p1} presents maximum accuracy based on the different number of devices ($S_c$) in each round in P1. We can observe that higher maximum accuracy can be obtained by increasing the value of $S_c$. For example, we can obtain $5.06\%$ maximum accuracy improvements as we increase $S_c$ from $25$ to $50$.
However, as we demonstrated in the part of communication efficiency analysis, increasing $S_c$ inevitably increases the same multiple in communication burden.

\section{CONCLUSION}\label{sec:conclu}
In this paper, we presented a novel FL method named CyclicFL,  which 
enables quick 
pre-training for effective 
initial models on AIoT devices using task-specific data instead of relying on unavailable public proxy data.
We analyzed and showed that the data consistency between the pre-training and federated training is significant to the success of our method.
Furthermore, we proved that our method can achieve faster convergence speed on diverse convexity scenarios.
Since the pre-trained model can provide flatter loss landscapes to guide the SGD processes during the FL training, 
the overall FL training process can be accelerated. We showed that our method can guarantee a flatter loss landscape from the perspective of data consistency.
The experimental results demonstrated the superiority of our method in speeding up convergence and improving classification accuracy.
Comprehensive experimental results  show 
the effectiveness of our proposed CyclicFL.

\bibliographystyle{plain}
\bibliography{mm}

\begin{thebibliography}{10}

\bibitem{iot_aiot}
Power consumption reduction for iot devices thanks to edge-ai: Application to human activity recognition.
\newblock 24:100930, 2023.

\bibitem{Caldas-et-al:leaf}
Sebastian Caldas, Sai Meher~Karthik Duddu, Peter Wu, Tian Li, Jakub Kone{\v{c}}n{\`y}, H~Brendan McMahan, Virginia Smith, and Ameet Talwalkar.
\newblock Leaf: A benchmark for federated settings.
\newblock {\em arXiv preprint arXiv:1812.01097}, 2018.

\bibitem{Chen-et-al:prefl}
Hong-You Chen, Cheng-Hao Tu, Ziwei Li, Han-Wei Shen, and Wei-Lun Chao.
\newblock On pre-training for federated learning.
\newblock {\em arXiv preprint arXiv:2206.11488}, 2022.

\bibitem{Chen-et-al:fedtune}
Jinyu Chen, Wenchao Xu, Song Guo, Junxiao Wang, Jie Zhang, and Haozhao Wang.
\newblock Fedtune: A deep dive into efficient federated fine-tuning with pre-trained transformers.
\newblock {\em arXiv preprint arXiv:2211.08025}, 2022.

\bibitem{Gao-et-al:gao2022feddc}
Liang Gao, Huazhu Fu, Li~Li, Yingwen Chen, Ming Xu, and Cheng-Zhong Xu.
\newblock Feddc: Federated learning with non-iid data via local drift decoupling and correction.
\newblock In {\em Proceedings of the IEEE/CVF Conference on Computer Vision and Pattern Recognition (CVPR)}, pages 10112--10121, 2022.

\bibitem{Glorot-Bengio:Xavier}
Xavier Glorot and Yoshua Bengio.
\newblock Understanding the difficulty of training deep feedforward neural networks.
\newblock In {\em Proceedings of the International Conference on Artificial Intelligence and Statistics (AISTATS)}, pages 249--256, 2010.

\bibitem{He-et-al:Heinit}
Kaiming He, Xiangyu Zhang, Shaoqing Ren, and Jian Sun.
\newblock Delving deep into rectifiers: Surpassing human-level performance on imagenet classification.
\newblock In {\em Proceedings of the IEEE International Conference on Computer Vision (ICCV)}, pages 1026--1034, 2015.

\bibitem{He-et-al:he2016deep}
Kaiming He, Xiangyu Zhang, Shaoqing Ren, and Jian Sun.
\newblock Deep residual learning for image recognition.
\newblock In {\em Proceedings of the IEEE Conference on Computer Vision and Pattern Recognition (CVPR)}, pages 770--778, 2016.

\bibitem{Hinton-et-al:distill}
Geoffrey Hinton, Oriol Vinyals, Jeff Dean, et~al.
\newblock Distilling the knowledge in a neural network.
\newblock {\em arXiv preprint arXiv:1503.02531}, 2(7), 2015.

\bibitem{Howard-et-al:howard2017mobilenets}
Andrew~G Howard, Menglong Zhu, Bo~Chen, Dmitry Kalenichenko, Weijun Wang, Tobias Weyand, Marco Andreetto, and Hartwig Adam.
\newblock Mobilenets: Efficient convolutional neural networks for mobile vision applications.
\newblock {\em arXiv preprint arXiv:1704.04861}, 2017.

\bibitem{Howard-et-al:Fintune}
Jeremy Howard and Sebastian Ruder.
\newblock Universal language model fine-tuning for text classification.
\newblock In {\em Proceedings of the Annual Meeting of the Association for Computational Linguistics (ACL)}, pages 328--339, 2018.

\bibitem{Hu-et-al:Hu2020Provable}
Wei Hu, Lechao Xiao, and Jeffrey Pennington.
\newblock Provable benefit of orthogonal initialization in optimizing deep linear networks.
\newblock In {\em Proceedings of the International Conference on Learning Representations (ICLR)}, 2020.

\bibitem{Karimireddy-et-al:scaffold}
Sai~Praneeth Karimireddy, Satyen Kale, Mehryar Mohri, Sashank Reddi, Sebastian Stich, and Ananda~Theertha Suresh.
\newblock Scaffold: Stochastic controlled averaging for federated learning.
\newblock In {\em Proceedings of the International Conference on Machine Learning (ICML)}, pages 5132--5143, 2020.

\bibitem{Krizhevsky-et-al:cifar}
Alex Krizhevsky, Geoffrey Hinton, et~al.
\newblock Learning multiple layers of features from tiny images.
\newblock 2009.

\bibitem{LeCun-et-al:Lenet}
Yann LeCun, L{\'e}on Bottou, Yoshua Bengio, and Patrick Haffner.
\newblock Gradient-based learning applied to document recognition.
\newblock {\em Proceedings of the IEEE}, 86(11):2278--2324, 1998.

\bibitem{Li-et-al:lossland}
Hao Li, Zheng Xu, Gavin Taylor, Christoph Studer, and Tom Goldstein.
\newblock Visualizing the loss landscape of neural nets.
\newblock In {\em Proceedings of the Advances in Neural Information Processing Systems (NeurIPS)}, volume~31, 2018.

\bibitem{Li-et-al:moon}
Qinbin Li, Bingsheng He, and Dawn Song.
\newblock Model-contrastive federated learning.
\newblock In {\em Proceedings of the IEEE/CVF Conference on Computer Vision and Pattern Recognition (CVPR)}, pages 10713--10722, 2021.

\bibitem{Li-et-al:fedproxy}
Tian Li, Anit~Kumar Sahu, Manzil Zaheer, Maziar Sanjabi, Ameet Talwalkar, and Virginia Smith.
\newblock Federated optimization in heterogeneous networks.
\newblock In {\em Proceedings of Machine Learning and Systems (MLSys))}, volume~2, pages 429--450, 2020.

\bibitem{Li-et-al:fedprox}
Tian Li, Anit~Kumar Sahu, Manzil Zaheer, Maziar Sanjabi, Ameet Talwalkar, and Virginia Smith.
\newblock Federated optimization in heterogeneous networks.
\newblock {\em Proceedings of Machine Learning and Systems (MLSys)}, pages 429--450, 2020.

\bibitem{li2023convergence}
Yipeng Li and Xinchen Lyu.
\newblock Convergence analysis of sequential federated learning on heterogeneous data.
\newblock In {\em Thirty-seventh Conference on Neural Information Processing Systems ((NeurIPS))}, 2023.

\bibitem{Liu-et-al:transferunder}
Hong Liu, Mingsheng Long, Jianmin Wang, and Michael~I Jordan.
\newblock Towards understanding the transferability of deep representations.
\newblock {\em arXiv preprint arXiv:1909.12031}, 2019.

\bibitem{lu2023federatedmm}
Jianghu Lu, Shikun Li, Kexin Bao, Pengju Wang, Zhenxing Qian, and Shiming Ge.
\newblock Federated learning with label-masking distillation.
\newblock In {\em Proceedings of the 31st ACM International Conference on Multimedia}, pages 222--232, 2023.

\bibitem{fediot}
Zili Lu, Heng Pan, Yueyue Dai, Xueming Si, and Yan Zhang.
\newblock Federated learning with non-iid data: A survey.
\newblock In {\em IEEE Internet of Things Journal}, volume~11, pages 19188--19209, 2024.

\bibitem{Mao-et-al:neucam}
Haitao Mao, Xu~Chen, Qiang Fu, Lun Du, Shi Han, and Dongmei Zhang.
\newblock Neuron campaign for initialization guided by information bottleneck theory.
\newblock In {\em Proceedings of the ACM International Conference on Information \& Knowledge Management (CIKM)}, page 3328–3332, 2021.

\bibitem{McMahan-et-al:fedavg}
Brendan McMahan, Eider Moore, Daniel Ramage, Seth Hampson, and Blaise~Aguera y~Arcas.
\newblock Communication-efficient learning of deep networks from decentralized data.
\newblock In {\em Proceedings of the Artificial Intelligence and Statistics (AISTATS)}, pages 1273--1282, 2017.

\bibitem{Nair-Hinton:rectified}
Vinod Nair and Geoffrey~E Hinton.
\newblock Rectified linear units improve restricted boltzmann machines.
\newblock In {\em Proceedings of the International Conference on Machine Learning (ICML)}, 2010.

\bibitem{Nguyen-et-al:wheretobegin}
John Nguyen, Kshitiz Malik, Maziar Sanjabi, and Michael Rabbat.
\newblock Where to begin? exploring the impact of pre-training and initialization in federated learning.
\newblock {\em arXiv preprint arXiv:2206.15387}, 2022.

\bibitem{Nwankpa-et-al:sigtanh}
Chigozie Nwankpa, Winifred Ijomah, Anthony Gachagan, and Stephen Marshall.
\newblock Activation functions: Comparison of trends in practice and research for deep learning.
\newblock {\em arXiv preprint arXiv:1811.03378}, 2018.

\bibitem{Radford-et-al:CLIP}
Alec Radford, Jong~Wook Kim, Chris Hallacy, Aditya Ramesh, Gabriel Goh, Sandhini Agarwal, Girish Sastry, Amanda Askell, Pamela Mishkin, Jack Clark, et~al.
\newblock Learning transferable visual models from natural language supervision.
\newblock In {\em Proceedings of the International Conference on Machine Learning (ICML)}, pages 8748--8763, 2021.

\bibitem{Savitzky-et-al:filter}
Abraham Savitzky and Marcel~JE Golay.
\newblock Smoothing and differentiation of data by simplified least squares procedures.
\newblock {\em Analytical Chemistry}, 36(8):1627--1639, 1964.

\bibitem{Simonyan-Zisserman:verydeep}
Karen Simonyan and Andrew Zisserman.
\newblock Very deep convolutional networks for large-scale image recognition.
\newblock {\em arXiv preprint arXiv:1409.1556}, 2014.

\bibitem{Wang-et-al:fedNova}
Jianyu Wang, Qinghua Liu, Hao Liang, Gauri Joshi, and H~Vincent Poor.
\newblock Tackling the objective inconsistency problem in heterogeneous federated optimization.
\newblock In {\em Proceedings of the Advances in neural information processing systems (NeurIPS)}, pages 7611--7623, 2020.

\bibitem{Xiao-et-al:fashmnist}
Han Xiao, Kashif Rasul, and Roland Vollgraf.
\newblock Fashion-mnist: a novel image dataset for benchmarking machine learning algorithms.
\newblock {\em arXiv preprint arXiv:1708.07747}, 2017.

\bibitem{Zhu-et-al:zhu2021gradinit}
Chen Zhu, Renkun Ni, Zheng Xu, Kezhi Kong, W~Ronny Huang, and Tom Goldstein.
\newblock Gradinit: Learning to initialize neural networks for stable and efficient training.
\newblock In {\em Proceedings of the Advances in Neural Information Processing Systems (NeurIPS)}, volume~34, pages 16410--16422, 2021.

\bibitem{Zhu-et-al:noniidsurvey}
Hangyu Zhu, Jinjin Xu, Shiqing Liu, and Yaochu Jin.
\newblock Federated learning on non-iid data: A survey.
\newblock {\em Neurocomputing}, pages 371--390, 2021.

\end{thebibliography}

\end{document}


\sloppy

\def\x{{\mathbf x}}
\def\L{{\cal L}}

\begin{table*}
\caption{Comparison of test accuracy. 
}
\label{exp_randomseeed}
\centering
\scalebox{0.8}{
\begin{tabular}{c|l|l|c|ccccc} 
\hline
\multirow{2}{*}{Model} & \multicolumn{2}{c|}{\multirow{2}{*}{Dataset}} & \multicolumn{1}{l|}{\multirow{2}{*}{\begin{tabular}[c]{@{}l@{}}Non-IID \\ Setting ($\beta$)\end{tabular}}} & \multicolumn{5}{c}{Accuracy (\%)} \\ 
\cline{5-9}
 & \multicolumn{2}{c|}{} & \multicolumn{1}{l|}{} & FedAvg & FedProx & SCAFFOLD & Moon & Cyclic + FedAvg \\ 
\hline\hline
CNN-FEMNIST & \multicolumn{2}{c|}{FEMNIST} & - & $81.44\pm1.26$  & $81.44\pm1.11$  & $\mathbf{83.48\pm0.48}$  & $81.46\pm0.90$  & $82.95\pm0.40$  \\ 
\hline
\multirow{3}{*}{CNN-Fashion-MNIST} & \multicolumn{2}{c|}{\multirow{3}{*}{Fashion-MNIST}} & 0.1 & $76.18\pm1.43$ & $76.11\pm2.01$ & $78.50\pm2.21$ & $76.28\pm1.34$ & $\mathbf{83.11\pm1.09}$ \\
 & \multicolumn{2}{c|}{} & 0.5 & $84.02\pm2.16$ & $84.01\pm2.19$ & $85.37\pm0.93$ & $83.94\pm1.97$ & $\mathbf{86.83\pm0.62}$ \\
 & \multicolumn{2}{c|}{} & 1.0 & $86.28\pm1.46$ & $86.18\pm1.42$ & $86.11\pm1.39$ & $85.89\pm1.50$ & $\mathbf{88.46\pm0.23}$ \\ 
\hline
\multirow{3}{*}{LeNet-5} & \multicolumn{2}{c|}{\multirow{3}{*}{CIFAR-10}} & 0.1 & $49.05\pm0.84$ & $49.00\pm0.23$ & $\mathbf{53.40\pm0.86}$ & $48.90\pm 0.45$ & $53.38\pm0.71$ \\
 & \multicolumn{2}{c|}{} & 0.5 & $53.22\pm0.60$ & $52.88\pm1.38$ & $\mathbf{60.14\pm1.53}$ & $52.27\pm0.27$ & $57.91\pm0.04$ \\
 & \multicolumn{2}{c|}{} & 1.0 & $55.11\pm1.03$ & $55.01\pm1.40$ & $\mathbf{61.40\pm1.20}$ & $54.82\pm0.50$ & $57.46\pm1.00$ \\ 
\hline
\multirow{6}{*}{ResNet-8} & \multirow{6}{*}{CIFAR-100} & \multirow{3}{*}{Coarse} & 0.1 & $37.73\pm0.66$ & $38.71\pm0.66$ & $35.83\pm1.90$ & $37.09\pm0.17$ & $\mathbf{53.94\pm1.10}$ \\
 &  &  & 0.5 & $54.83\pm1.02$ & $54.72\pm0.01$ & $52.55\pm0.02$ & $54.28\pm0.73$ & $\mathbf{64.46\pm0.12}$ \\
 &  &  & 1.0 & $58.07\pm1.28$ & $57.66\pm0.57$ & $55.91\pm0.04$ & $57.63\pm0.37$ & $\mathbf{66.12\pm0.19}$ \\ 
\cline{3-9}
 &  & \multicolumn{1}{c|}{\multirow{3}{*}{Fine}} & 0.1 & $40.40\pm0.94$ & $40.80\pm0.14$ & $42.54\pm0.78$ & $41.93\pm0.49$ & $\mathbf{54.27\pm0.41}$ \\
 &  & \multicolumn{1}{c|}{} & 0.5 & $47.65\pm0.31$ & $47.79\pm0.68$ & $48.76\pm0.23$ & $47.76\pm0.24$ & $\mathbf{57.06\pm0.66}$ \\
 &  & \multicolumn{1}{c|}{} & 1.0 & $47.63\pm0.24$ & $48.54\pm0.16$ & $48.75\pm0.10$ & $48.76\pm0.11$ & $\mathbf{56.47\pm0.31}$ \\
\hline
\end{tabular}
}
\end{table*}

\section{More Experiments}\label{appen-exp}

Table~\ref{exp_randomseeed} shows the top-1 test accuracy and variances for all FL baselines under Non-IID settings.
We can observe that Cyclic+FedAvg greatly beat four basic FL algorithms for the CIFAR-100-Coarse/Fine dataset. For CIFAR-100-Coarse with $\beta=0.5$, Cyclic+FedAvg outperforms FedAvg, FedProx, SCAFFOLD, and Moon by $9.63\%$, $9.74\%$, $11.91\%$, and $10.18\%$, respectively.
For the Fashion-MNIST dataset, the cyclic training also consistently outperforms all baselines.
For CIFAR-10 and FEMNIST, we observe that SCAFFOLD consistently outperforms other methods, including Cyclic+FedAvg. Since our method is a pre-training process, the higher classification accuracy of SCAFFOLD compared to Cyclic+FedAvg is not against the effectiveness of cyclic training.

\section{More convergence analyses}\label{more convergence}
In this section, we presented the convergence analyses for strongly convex and general convex scenarios. We extended the analyses in \cite{li2023convergence} and showed the convergence rate of our proposed method.
\begin{assumption} \label{assum4}
There exist a constant $\zeta_*^2$ such that
\begin{align}
    \nonumber\frac{1}{m}\sum_{i=1}^m\left\|\nabla L_i(\mathbf{x}^*)\right\|^2=\zeta_*^2
\end{align}
where $\mathbf{x}^*$ is the global minimizer.
\end{assumption}

\begin{corollary}\textbf{(Strongly convex)}
Let $\mathbb{E}$ represent 
$\mathbb{E}\left[\mathcal{L}(\bar{\mathbf{x}}^{T})-\mathcal{L}(\mathbf{x}^{*})\right]$.
Let $D_c$ represent $\|\mathbf{x}^0 - \mathbf{x}^{T_c} \|$ and $D_p$ represent $\|\mathbf{x}^{T_c} - \mathbf{x}^{T_{total}} \|$.
Under \textbf{Assumptions~\ref{assum1},\ref{assum2}, \ref{assum4}}, and \textbf{Definition~\ref{def}}, there exists a constant effective learning rate $\frac{\mu}{T}\leq\tilde{\eta}\leq\frac{1}{6L}$ that the convergence rate in a strongly convex scenario is defined as

\tiny
\noindent for $T\leq T_c$ ,
\begin{align}
\nonumber\mathbb{E}=\mathcal{O}\left(\mu D_c^{2}\exp\left(-\frac{\mu R}{12L}\right)+\frac{\sigma^{2}}{\mu S_cK_cT}+
\frac{(m-S_c)\zeta_{*}^{2}}{\mu S_cT(m-1)}+\frac{L\sigma^{2}}{\mu^{2}S_cK_cT^{2}}+\frac{L\zeta_{*}^{2}}{\mu^{2}S_cT^{2}}\right)
;
\end{align}

\noindent for $T_c<T<T_{total}$ ,
\begin{align}
\nonumber\mathbb{E}=\mathcal{O}\left(\mu D_p^{2}\exp\left(-\frac{\mu T}{12L}\right)+\frac{\sigma^{2}}{\mu S_pK_pT}+\frac{(m-S_p)\zeta_{*}^{2}}{\mu S_pT(m-1)}+\frac{L\sigma^{2}}{\mu^{2}K_pT^{2}}+\frac{L\zeta_{*}^{2}}{\mu^{2}T^{2}}\right).
\end{align}

\end{corollary}

\begin{corollary}\textbf{(General convex)}
Let $\mathbb{E}$ represent 
$\mathbb{E}\left[\mathcal{L}(\bar{\mathbf{x}}^{T})-\mathcal{L}(\mathbf{x}^{*})\right]$. Let $D_c$ represent $\|\mathbf{x}^0 - \mathbf{x}^{T_c} \|$ and $D_p$ represent $\|\mathbf{x}^{T_c} - \mathbf{x}^{T_{total}} \|$
Under \textbf{Assumptions~\ref{assum1},\ref{assum2}, \ref{assum4}}, and \textbf{Definition~\ref{def}}, there exists a constant effective learning rate $\tilde{\eta}\leq\frac{1}{6L}$ that the convergence rate in a general convex scenario is defined as

\tiny
\noindent for $T\leq T_c$ ,
\begin{align}
\nonumber\mathbb{E}=
\mathcal{O}\left(\frac{\sigma D_c}{\sqrt{S_cK_c}}+\sqrt{1-\frac{S_c}{M_c}}\cdot\frac{\zeta_{*}D_c}{\sqrt{S_cT}}+\frac{\left(L\sigma^{2}D_c^{4}\right)^{1/3}}{(S_cK_c)^{1/3}T^{2/3}}+\frac{\left(L\zeta_{*}^{2}D_c^{4}\right)^{1/3}}{S_c^{1/3}T^{2/3}}+\frac{LD_c^{2}}{T}\right)
;
\end{align}

\noindent for $T_c<T<T_{total}$ ,
\begin{align}
\scriptsize
\nonumber\mathbb{E}=\mathcal{O}\left(\frac{\sigma D_p}{\sqrt{S_pK_pT}}+\sqrt{1-\frac{S_p}{m}}\cdot\frac{\zeta_{*}D_p}{\sqrt{S_pT}}+\frac{\left(L\sigma^{2}D_p^{4}\right)^{1/3}}{K_p^{1/3}T^{2/3}}+\frac{\left(L\zeta_{*}^{2}D_p^{4}\right)^{1/3}}{T^{2/3}}+\frac{LD_p^{2}}{T}\right).
\end{align}

\end{corollary}